\documentclass[11pt,letterpaper]{article}

\usepackage[margin=1in]{geometry}
\usepackage{amsmath}

\usepackage[round]{natbib}

\usepackage{amsthm, amsmath, amssymb}
\usepackage{xcolor}

\newtheoremstyle{customremark}
  {\topsep}
  {\topsep}
  {\normalfont}
  {0pt}
  {\bfseries}
  {.}
  { }
  {\thmname{#1}\thmnumber{ #2}\thmnote{ (#3)}}

\theoremstyle{customremark}
\newtheorem{rem}{Remark}[section]

\theoremstyle{plain}
\newtheorem{lemma}{Lemma}[section]
\newtheorem{proposition}{Proposition}
\newtheorem{assumption}{Assumption}[section]
\counterwithout{assumption}{section}

\newtheorem{theorem}{Theorem}
\newtheorem{corollary}{Corollary}[section]

\newtheorem{definition}{Definition}[section]

\usepackage{mathrsfs,mathtools,booktabs,array,longtable}
\usepackage{hyperref}
\providecommand{\Ent}{\operatorname{Ent}}
\providecommand{\Var}{\operatorname{Var}}
\providecommand{\Cov}{\operatorname{Cov}}

\providecommand{\op}{\mathrm{op}}
\providecommand{\bil}{\mathrm{bil}}
\providecommand{\HS}{\mathrm{HS}}
\providecommand{\CP}{C_{\mathrm P}}
\providecommand{\CLS}{C_{\mathrm{LS}}}
\providecommand{\calL}{\mathscr L}
\providecommand{\V}{\mathcal V}
\providecommand{\E}{\mathbb E}
\providecommand{\R}{\mathbb R}
\providecommand{\norm}[1]{\left\lVert#1\right\rVert}

\hypersetup{hidelinks}

\usepackage[dvipsnames]{xcolor}
\usepackage[]{color-edits}
\addauthor{ab}{violet}

\title{Width Improves Posterior Sampling in Bayesian Neural Networks}
\author{%
  Nalin Tiwary\thanks{Equal Contribution}
  \qquad Meenal Jhajharia\footnotemark[1]
  \qquad Arindam Banerjee\\
  University of Illinois Urbana-Champaign
}
\date{}

\begin{document}

\maketitle

\begin{abstract}
Increasing neural network width has been demonstrated to simplify optimization, but its effect on  Bayesian neural network posterior sampling is less understood. We show that width improves the effectiveness of posterior sampling for classification networks with bounded smooth activations and Gaussian priors, under explicit bounds on the prior means. For a network of width $m$ and prior variance $\sigma^2$, the unrestricted posterior has Poincar\'e constant at most $\sigma^2+O(m^{-1/3})$ at fixed data and depth. The global Poincar\'e bound follows from controlling negative curvature and removing weight restrictions. Exploiting convexity in the output weights yields sharper logarithmic Sobolev inequalities (LSI). With two layers and $n$ observations, the global LSI constant is at most $\sigma^2+O(n/\sqrt m+n^2/m)$, approaching the prior scale when $m\gg n^2$. For general fixed depth, when $m\gg n^3$, the same conclusion holds on a convex spectral domain with posterior mass tending to one, leaving the output weights and first layer unrestricted. Experiments further support the predicted improvement in sampling with width. These results show that width based overparameterization can improve the geometry governing posterior fluctuations and Langevin relaxation, even as the parameter space grows.
\end{abstract}

\section{Introduction}
\label{sec:introduction}

Increasing neural network width can simplify optimization: in overparameterized regimes, gradient-based methods admit convergence guarantees despite the nonconvexity of the training objective \citep{jacot2018neural,du2019gradient}. Bayesian learning raises a related computational question. Predictions average over networks according to their posterior probabilities, making representative posterior samples central to this approach. Increasing width, however, also increases the number of weights that a sampler must explore. Can the same increase in width that helps optimization also make the Bayesian posterior easier to explore?

Representative sampling requires control of how probability is distributed across parameter space. Convergence to a favorable training solution alone provides little information about the barriers a sampler may encounter when moving between regions of substantial posterior mass. We study quantitative guarantees at finite width in the network's existing, width-normalized weight coordinates. The challenge is to control the nonconvex geometry created by the likelihood's nonlinear dependence on the weights.

We quantify this geometry through Poincar\'e and logarithmic Sobolev inequalities (PI and LSIs). A PI compares the variance of an observable with its sensitivity to changes in the weights. A large constant can reflect a probability barrier: an observable can differ across regions of substantial mass while changing mainly in a region of low probability. An LSI gives stronger entropy control and implies exponential decay of Kullback--Leibler divergence along continuous-time Langevin dynamics \citep[Theorem~1]{vempala2019rapid}. In our convention, smaller constants give stronger control and faster corresponding relaxation rates. An independent Gaussian prior with variance $\sigma^2$ in each coordinate provides a natural benchmark: both constants equal $\sigma^2$, regardless of the number of parameters \citep[Sections~4.1 and~5.1]{bakry2014analysis}.

We show that width can bring posterior bounds on these constants arbitrarily close to the Gaussian prior benchmark. Our setting is fully connected classification networks with bounded smooth activations, controlled activation derivatives, and independent Gaussian priors under explicit bounds that allow nonzero prior means. Write $m$ for the width of each hidden layer and $n$ for the number of observations.

Our first result controls the full posterior of a deep network: at fixed dataset and depth, its Poincar\'e constant is at most $\sigma^2+O(m^{-1/3})$ (Theorem~\ref{thm:global_pi}). All weights remain unrestricted in this conclusion. Thus, even as the parameter space grows, the excess in the variance bound over the Gaussian benchmark vanishes. The proof first controls nonlinear interactions on a restricted domain, then uses the Gaussian prior's restoring drift to extend the inequality to the entire parameter space. The stronger question is whether width also yields entropy control.

For a two-layer network, with one hidden layer, the full posterior has LSI constant at most $\sigma^2+O(n/\sqrt m+n^2/m)$ under a uniform bound on the input norms (Theorem~\ref{thm:shallow_lsi}). In particular, the sufficient regime $m\gg n^2$ brings the bound to the prior scale while allowing the sample size to grow. The key structure is that, with the hidden features fixed, the classification loss is convex in the output weights. Integrating over these weights and controlling their dependence on the hidden weights yields the stronger inequality. The required width also reflects the geometry of the inputs: for orthonormal data, the sufficient regime improves to $m\gg n$ (Remark~\ref{rem:lsi_input_geometry}).

At greater depth, a weight perturbation propagates through subsequent hidden matrices and can be amplified. We control this amplification on a convex domain that bounds the normalized operator norms of the later hidden matrices, leaving the first layer and output weights unrestricted. For the posterior conditioned on this domain, the LSI constant approaches $\sigma^2$ when $m\gg n^3$, at fixed depth and spectral cutoff (Theorem~\ref{thm:deep_lsi}). The same domain has posterior mass tending to one in this regime (Lemma~\ref{lem:spectral_mass}), so the stronger entropy control applies to a region capturing asymptotically all posterior probability. Experiments provide complementary support through improved decorrelation of sampled losses and predictions as width increases (Section~\ref{sec:experiments}). Together, the results show how width can improve the geometry governing posterior fluctuations and relaxation even as it enlarges the parameter space.

\paragraph{Related Work}
\label{sec:related_work}
Width can simplify Bayesian neural network posteriors in function space and under suitable changes of coordinates. \citet[Proposition~1]{hron2020exact} establish convergence of function-space posteriors to the posterior induced by the Gaussian-process limit. For Gaussian regression, \citet[Theorem~2.1]{hron2022wide} construct a data-dependent reparameterization under which the weight posterior approaches a Gaussian reference. Building on this representation, \citet[Theorems~2.1 and~2.2]{pezzetti2025function} show that acceptance probabilities for preconditioned Crank--Nicolson samplers and their Langevin variant tend to one with width, at fixed admissible step sizes. Our results complement these asymptotic descriptions by bounding functional inequalities for finite-width classification posteriors in the original coordinates, with explicit dependence on width and sample size.

A complementary line of work studies noisy neural-network training through mean-field Langevin dynamics. \citet[Theorem~1]{nitanda2022convex} and \citet[Theorem~3.2]{chizat2022meanfield} establish exponential convergence of entropy-regularized mean-field objectives under uniform LSI assumptions. This theory also reaches finite-particle systems: uniform-in-particle-number LSI bounds have been established for their stationary Gibbs measures (see \citealp[Corollary~23]{kook2024sampling}; \citealp[Theorem~1]{wang2024uniform}; \citealp{chewi2026uniform}). These results concern energies defined through the empirical distribution of particles. In their neural-network applications, neurons contribute through a mean-field average. Our setting uses inverse-square-root width normalization and a fixed Gaussian prior variance on individual weights.

Functional inequalities connect these questions of posterior geometry to quantitative sampling guarantees. \citet[Theorem~2]{vempala2019rapid} analyze the unadjusted Langevin algorithm under an LSI and a Lipschitz gradient assumption. \citet[Theorems~7 and~8]{chewi2025analysis} extend this analysis to weaker functional inequalities and H\"older-continuous gradients. These analyses express algorithmic convergence in terms of functional-inequality constants; we bound the constants for neural-network posteriors.

Our proof tools include two-scale criteria that combine conditional and marginal LSIs under quantitative control of their dependence \citep[Theorem~3]{grunewald2009twoscale}; \citet[Theorem~3.1, part~3]{srinivasan2026twoscale} gives a criterion using exponential-moment bounds for derivatives of the conditional log-density. The global PI argument uses the connection between local inequalities and Lyapunov conditions developed by \citet[Section~3.2]{bakry2008lyapunov}.

\section{Setup and Preliminaries}
\label{sec:setup}
\subsection{Model and posterior}
\label{sec:model}

We observe classification data $(x_i,y_i)_{i=1}^n$, where $x_i\in\R^d$ and $y_i\in\{1,\ldots,C\}$, with $C\ge2$. Consider a fully connected network without biases, with $L\ge2$ weight layers and width $m$ in every hidden layer. Its hidden weight matrices are $W_1\in\R^{m\times d}$ and $W_\ell\in\R^{m\times m}$ for $2\le\ell\le L-1$. The output matrix $A\in\R^{(C-1)\times m}$ is the head, and $\theta$ is the vector of all parameter entries.

Applying the activation $\phi:\R\to\R$ coordinatewise and using the standard inverse-square-root width normalization \citep{jacot2018neural,hron2022wide}, we define the hidden features and network output by
\begin{equation}
\label{eq:network}
\begin{aligned}
H_1(x)
    &= \phi(W_1x),\\
H_\ell(x)
    &= \phi\!\left(W_\ell H_{\ell-1}(x)/\sqrt m\right),\\
f_\theta(x)
    &= A H_{L-1}(x)/\sqrt m,
\end{aligned}
\end{equation}
The hidden-layer recursion applies for $2\le\ell\le L-1$.

Adding the same constant to every logit leaves softmax unchanged. We therefore represent the centered class logits as $Qf_\theta(x)$, where the columns of $Q\in\R^{C\times(C-1)}$ form an orthonormal basis of the zero-sum subspace of $\R^C$. Let $\ell_{\mathrm{CE}}$ denote the multiclass cross-entropy. The summed empirical loss is
\begin{equation}
\label{eq:loss}
V(\theta)
    = \sum_{i=1}^n
      \ell_{\mathrm{CE}}\!\left(Qf_\theta(x_i),y_i\right).
\end{equation}
We place independent Gaussian priors with variance $\sigma^2>0$ on the parameter entries. Let $\theta_0$ denote the prior mean and $Z$ the posterior normalizing constant. The prior and posterior are
\begin{equation}
\label{eq:posterior}
\begin{aligned}
\gamma
    &= \mathcal N(\theta_0,\sigma^2 I),\\
\pi(d\theta)
    &= Z^{-1}e^{-V(\theta)}\gamma(d\theta).
\end{aligned}
\end{equation}
The blocks of $\theta_0$ are denoted by $W_{1,0},\ldots,W_{L-1,0},A_0$. 

\subsection{Functional inequalities}
\label{subsec:funcineq}

Let $\mu$ be a probability measure. We write $\CP(\mu)$ and $\CLS(\mu)$ for the optimal constants in the Poincar\'e and logarithmic Sobolev inequalities ~\citep[Chapters~4--5]{bakry2014analysis}, respectively.

\begin{definition}[Poincar\'e inequality]
\label{def:pi}
The Poincar\'e constant $\CP(\mu)$ is the least constant such that
\begin{equation}
\Var_\mu(g)\le \CP(\mu)\int\norm{\nabla g}^2\,d\mu
\label{eq:pi}
\end{equation}
for every smooth test function $g$.
\end{definition}

\begin{definition}[Logarithmic Sobolev inequality]
\label{def:lsi}
The logarithmic Sobolev constant $\CLS(\mu)$ is the least constant such that
\begin{equation}
\label{eq:lsi_convention}
\Ent_\mu(g^2)\le 2\CLS(\mu)\int\norm{\nabla g}^2\,d\mu
\end{equation}
for every smooth test function $g$, where
$$
\Ent_\mu(g^2)
=
\int g^2\log\!\left(\frac{g^2}{\int g^2\,d\mu}\right)\,d\mu.
$$
\end{definition}
For the Gaussian prior $\gamma$, both optimal constants equal $\sigma^2$.
Bounds approaching $\sigma^2$ therefore express Gaussian-scale control of posterior variance and entropy relative to gradient energy in the original parameter coordinates.

\begin{rem}
Adding a large constant makes a function's fluctuations small relative
to its mean, while leaving its variance and gradient unchanged. In the
entropy calculation, the terms proportional to the deviations cancel
because deviations from the mean average to zero. The main contribution
comes from the squared deviations, whose average is the variance. Thus
the entropy control in an LSI yields variance control in the form of a
PI, with
$\CP(\mu)\le\CLS(\mu)$~\citep[Section~5.1]{bakry2014analysis}.
\qed
\end{rem}

\subsection{Assumptions}
\label{sec:assumptions}
The activation bounds and prior mean bounds apply to the Global PI and sharper LSI results. The sharper LSI bounds also require a uniform input bound when the sample size varies.

\begin{assumption}[Activation]
\label{ass:activation}
The activation $\phi\in C^2(\R)$ satisfies, for a finite constant $c_\phi$,
$$
\begin{gathered}
\sup_{t\in\R}|\phi(t)|\le1,
\qquad
\sup_{t\in\R}|\phi'(t)|\le1,\\
\sup_{t\in\R}|\phi''(t)|\le c_\phi.
\end{gathered}
$$
\end{assumption}

\begin{rem}
Tanh and logistic sigmoid satisfy these bounds.
The sharper LSI results require only $C^2$ regularity,
whereas the global PI argument assumes
$\phi\in C^\infty(\mathbb R)$, satisfied by many modern activations such as GELU and sin. Although GELU does not satisfy Assumption ~\ref{ass:activation} due to being unbounded.
\qed
\end{rem}

\begin{assumption}[Prior means]
\label{ass:prior_means}
Fix the dataset and depth $L$. The prior mean is deterministic,
and there are finite constants $a_0,r_0$, independent of width,
such that
$$
\norm{A_0}_{\op}\le a_0,\qquad
\frac{\norm{W_{\ell,0}}_{\op}}{\sqrt m}\le r_0
\quad (2\le\ell\le L-1).
$$
Here $\norm{\cdot}_{\op}$ denotes the matrix operator norm.
The first-layer mean $W_{1,0}$ is arbitrary. For $L=2$, only
the bound on $A_0$ applies. For the sample-size-dependent LSI
results, these bounds hold with the same constants as $n$ varies.
\end{assumption}

\begin{rem}
Centered Gaussian priors, a standard choice in wide-network analysis \citep{hron2022wide}, satisfy these bounds. Together with bounded activation values, the bound on $A_0$ keeps the prior predictions at a width-independent scale. The permitted $\sqrt m$ scale for later hidden means matches the operator-norm scale of Gaussian matrix fluctuations
\citep{vershynin2012introduction}.
\qed
\end{rem}

\begin{assumption}[Bounded inputs]
\label{ass:inputs}
There is a constant $\bar{x}<\infty$, independent of $n,m$, such that
$$
\max_{1\le i\le n}\norm{x_i}\le \bar{x}.
$$
\end{assumption}
 
\begin{rem}
The uniform input bound is used for the sample-size-dependent
LSI estimates. The global PI result fixes the dataset and
allows its constants to depend on the input norms.
\qed
\end{rem}

\section{Global Poincar\'e inequality at fixed depth}
\label{sec:global_pi}
Our first result gives a global Poincar\'e bound that approaches the
Gaussian prior's constant as width increases at fixed depth.
Table~\ref{tab:pi_lsi_constants} summarizes the model quantities that
enter both the PI bounds here and the LSI bounds in Section~\ref{sec:lsi}.
\begin{table}[!ht]
\centering
\caption{Model quantities shared by the PI and LSI bounds.}
\label{tab:pi_lsi_constants}
\renewcommand{\arraystretch}{1.15}
\begin{tabular}{@{}l p{0.81\linewidth}@{}}
\hline
\textbf{Symbol} & \textbf{Meaning}\\
\hline
$C$ & Number of classes.\\
$\sigma$ & Prior standard deviation of each parameter entry.\\
$c_\phi$ & Upper bound on $\sup_t|\phi''(t)|$.\\
$a_0$ & Bound on $\norm{A_0}_{\op}$ for the output-weight prior mean $A_0$.\\
$r_0$ & Upper bound on $\norm{W_{\ell,0}/\sqrt m}_{\op}$ for $\ell=2,\ldots,L-1$.\\
\hline
\end{tabular}
\end{table}

\begin{theorem}[Global PI with explicit depth dependence]
\label{thm:global_pi}
Suppose Assumptions~\ref{ass:activation} and~\ref{ass:prior_means} hold,
with $\phi\in C^\infty(\mathbb R)$. Fix the dataset and the quantities
$C,\sigma,c_\phi,a_0,r_0$. For any fixed
$\alpha>\max\{1,r_0+3.853\sigma\}$, there exist $K_0,K>0$ depending
only on the dataset and these fixed quantities, including $\alpha$.
Then for every depth $L>2$ and width $m$, the full posterior $\pi$
in~\eqref{eq:posterior} satisfies
\begin{equation}
\label{eq:global_pi}
m\ge K_0\alpha^{9L}
\quad\Longrightarrow\quad
\CP(\pi)\le\sigma^2+K \alpha^{3L}m^{-1/3}.
\end{equation}
\end{theorem}

The width condition is sufficient for the bound in~\eqref{eq:global_pi}
with constants $K_0$ and $K$ independent of depth and width.

\paragraph{Proof sketch.}
We first prove a PI on a restricted domain. We then extend it to the full
posterior by removing the restrictions one at a time.

\paragraph{Step 1. Define the local domain and bound its posterior mass.}
The posterior potential is
$U(\theta)=V(\theta)+\norm{\theta-\theta_0}^2/(2\sigma^2)$.
Its Hessian contains the Gaussian curvature, a positive semidefinite
term inherited from cross-entropy in the logits, and a nonlinear-feature
term that can have either sign. Define
\begin{equation}
\label{eq:pi_curvature_deficit}
D(\theta)=\max\left\{0,\,
\sigma^{-2}-\lambda_{\min}\!\left(\nabla^2U(\theta)\right)
\right\},
\end{equation}
where $\lambda_{\min}$ denotes the smallest eigenvalue. Then
\[
\nabla^2U(\theta)\succeq
\bigl(\sigma^{-2}-D(\theta)\bigr)I.
\]
To control the nonlinear term, use the cutoff $\alpha$ to bound the
normalized operator norms of $W_2,\ldots,W_{L-1}$ and a radius $B>0$
to bound the displacement of each head column from its prior mean.
Write $A=[a_1,\ldots,a_m]$ and $A_0=[a_{1,0},\ldots,a_{m,0}]$. Define
\begin{equation}
\label{eq:pi_cylinder}
\begin{aligned}
\mathcal H_B
    &= \{\theta:\max_j\norm{a_j-a_{j,0}}\le B\},\\
\mathcal G_\alpha
    &= \bigcap_{\ell=2}^{L-1}
       \{\theta:\norm{W_\ell/\sqrt m}_{\op}\le \alpha\},\\
\mathcal G_{B,\alpha}
    &= \mathcal H_B\cap\mathcal G_\alpha.
\end{aligned}
\end{equation}
These restrictions preserve convexity and leave $W_1$ unrestricted,
since the activation already bounds the features entering subsequent
layers. Let $\nu=\pi(\cdot\mid\mathcal G_{B,\alpha})$.

Bounded features and the bound on $A_0$ give a width-independent bound
on $\E_\gamma V$. Jensen's inequality yields
$Z\ge\exp(-\E_\gamma V)$, and $V\ge0$ gives
$\pi(E)\le\gamma(E)/Z$ for every measurable event $E$.
Thus, at fixed depth, Gaussian column and operator-norm tails give
$\pi(\mathcal G_{B,\alpha})\to1$ when $B/\sqrt{\log m}\to\infty$
(Appendix Lemma~\ref{lem:mass}).

\paragraph{Step 2. Prove a PI on the local domain.}
On $\mathcal G_{B,\alpha}$, the derivative estimates reduce control of
$D(\theta)$ to control of head projections $Av$, where
$v\in\mathbb R^m$ depends on the data and hidden weights. Conditional
on the hidden weights under the prior, each projection is Gaussian
with mean $A_0v$ and covariance $\sigma^2\norm v^2 I$. The spectral
constraints bound $\norm v$ independently of width at fixed depth.
Gaussian concentration and a union bound over the relevant projections
give exponential-moment estimates with a $\sqrt{\log m}$ contribution
from the maximum (Appendix Lemmas~\ref{lem:hessian} and~\ref{lem:tail}).
These estimates transfer to $\nu$ using
\[
\nu(E)\le
\frac{\gamma(E\cap\mathcal G_{B,\alpha})}
     {Z\pi(\mathcal G_{B,\alpha})},
\]
whose denominator is bounded away from zero by Step~1.

The integrated Bochner identity~\eqref{eq:pi_bochner_sketch} in
Appendix~\ref{app:PI} relates the squared generator to a nonnegative
Hessian term and the curvature of $U$
\citep{bakry2014analysis,kolesnikov2017brascamp}.
Appendix Remark~\ref{rem:bochner_background} explains how the identity
gives a PI. The deficit in~\eqref{eq:pi_curvature_deficit} then requires control of
$\int D\norm{\nabla g}^2\,d\nu$. The Gaussian prior conditioned on the
convex domain satisfies an LSI with constant at most $\sigma^2$
\citep[Theorem~A1]{schlichting2019mixtures}. The cutoffs bound
$\norm{\nabla V}$, allowing this inequality to transfer to an entropy
bound for $\nu$ with an additive multiple of $\int g^2\,d\nu$
(Appendix Lemma~\ref{lem:dlsi}). Entropy duality then lets us combine the
exponential moments of $D$ with this bound applied to
$\norm{\nabla g}$, regularized at zero. Using
$\norm{\nabla\norm{\nabla g}}\le\norm{\nabla^2g}_{\HS}$ gives
\begin{equation}
\label{eq:pi_weighted_curvature}
\begin{aligned}
\int D\norm{\nabla g}^2\,d\nu
\le{}&\frac12\int\norm{\nabla^2g}_{\HS}^2\,d\nu\\
     &+\Delta\int\norm{\nabla g}^2\,d\nu.
\end{aligned}
\end{equation}
At fixed data, depth, and cutoff, the error satisfies
\[
\Delta=O_{L,\alpha}\!\left(
\frac{B+1+\sqrt{\log m}}{\sqrt m}+\frac{1}{m}\right).
\]
On smooth convex approximations to the domain, the positive Hessian term
in the Bochner identity absorbs the first term in
\eqref{eq:pi_weighted_curvature}, and convexity makes the boundary term
nonnegative.
The resulting local bound is
$\CP(\nu)\le(\sigma^{-2}-\Delta)^{-1}$ when $\Delta<\sigma^{-2}$.
Theorems~\ref{thm:transfer} and~\ref{thm:local} in Appendix~\ref{app:PI}
give the details, including passage to $\mathcal G_{B,\alpha}$.

\paragraph{Step 3. Remove the hidden-layer restrictions.}
Release $W_2,\ldots,W_{L-1}$ in increasing layer order while retaining
$\mathcal H_B$. Bounded features from the earlier layers and the
remaining spectral bounds on the later layers control the likelihood
gradient for the matrix being released. For $B=o(\sqrt m)$ at fixed
depth, the Gaussian restoring drift dominates outside the cutoff.
Each application of the gluing lemma adds $O_{L,\alpha}(m^{-1})$
to the Poincar\'e bound and multiplies the preceding bound by
$1+O_{L,\alpha}(e^{-cm})$ for some $c>0$
(Appendix Lemmas~\ref{lem:net} and~\ref{lem:glue}).
See also \citet{bakry2008lyapunov} for this use of inward drift.

\paragraph{Step 4. Remove the head restriction.}
Only $\mathcal H_B$ remains after the hidden layers have been released.
Bounded final features give $\norm{\nabla_{a_j}V}\le 2n/\sqrt m$
for each head column on the full parameter space.
The Gaussian drift grows with the column's displacement from its
prior mean and controls the last application of the gluing lemma.
For $B/\sqrt{\log m}\to\infty$, this extension costs $O(B^{-2})$
(Theorem~\ref{thm:global} in Appendix~\ref{app:PI}).
with $B=o(\sqrt m)$ gives
\begin{equation}
\label{eq:pi_error_balance}
\begin{aligned}
\CP(\pi)\le\sigma^2+O_{L,\alpha}\Bigl(
    &\frac{B+1+\sqrt{\log m}}{\sqrt m}\\
    &+\frac1{B^2}+\frac1m\Bigr).
\end{aligned}
\end{equation}

\paragraph{Step 5. Choose the radius and track depth.}
Taking $B\asymp m^{1/6}$ balances the local curvature cost $B/\sqrt m$
with the head-extension cost $B^{-2}$. This radius gives cylinder mass
tending to one at fixed depth and satisfies $B=o(\sqrt m)$.
The constants in~\eqref{eq:pi_error_balance} may depend on $L$.
Appendix~\ref{app:global_pi_depth} tracks the depth dependence and uses
$m\ge K_0\alpha^{9L}$ to ensure a positive local spectral gap,
a lower bound on the cylinder mass, and the drift bounds needed to
remove the restrictions. This yields the error
$K\alpha^{3L}m^{-1/3}$ in~\eqref{eq:global_pi}.
\qed

\begin{rem}[Variance and entropy]
\label{rem:pi_lsi_globalization}
The gluing argument starts by subtracting the test function's mean under
the conditioned posterior. This leaves its variance under the full
posterior unchanged. The local PI controls the squared fluctuations
inside the restricted domain, and the inward drift controls the
contribution outside. Together these estimates give a global PI
(Appendix Lemma~\ref{lem:glue}).

An entropy bound also requires control of the logarithmic factor in
the contribution from outside the domain. Section~\ref{sec:lsi} uses
convexity in the output weights and an analysis of the hidden-weight
marginal to obtain a global LSI for $L=2$ and a local LSI for $L>2$.
\qed
\end{rem}

\begin{table*}[t]
\centering\small
\setlength{\tabcolsep}{4pt}
\caption{Main functional inequalities. The LSI bounds also imply PI
bounds. Here $\pi$ is the full posterior and
$\nu_a=\pi(\cdot\mid\mathcal G_a)$, with the hidden spectral domain
$\mathcal G_a$ defined in~\eqref{eq:lsi_domain}. All limits fix depth
and the activation and prior scales. The global PI fixes the dataset;
the LSI results assume uniformly bounded inputs as $n$ varies.
Constants and finite-width conditions are specified in the cited
theorems.\\}
\label{tab:main-results}
\begin{tabular}{@{}p{.11\textwidth}p{.14\textwidth}p{.43\textwidth}p{.25\textwidth}@{}}
\toprule
Depth & Posterior & Upper bound & Convergence regime\\
\midrule
$L>2$ & Full $\pi$
  & $\CP(\pi)\le\sigma^2+K \alpha^{3L}m^{-1/3}$
  & Fixed data, $m\to\infty$; Theorem~\ref{thm:global_pi}\\[5pt]
$L=2$ & Full $\pi$
  & $\CLS(\pi)\le\sigma^2+O(n/\sqrt m+n^2/m)$
  & $n^2/m\to0$; Theorem~\ref{thm:shallow_lsi}\\[5pt]
$L>2$ & Conditioned $\nu_a$
  & $\CLS(\nu_a)\le\sigma^2+O_{L,a}(n^{3/2}/\sqrt m+n^2/m)$
  & $n^3/m\to0$; Theorem~\ref{thm:deep_lsi}\\
\bottomrule
\end{tabular}
\par\smallskip
For the deep LSI, $\pi(\mathcal G_a^c)\le(L-2)e^{c_0n-c_1m}$
by Lemma~\ref{lem:spectral_mass}. The spectral cutoffs satisfy
$\alpha>\max\{1,r_0+3.853\sigma\}$ for the global PI proof and
$a>r_0+2\sigma$ for the deep LSI.
\end{table*}

\section{Sharper logarithmic Sobolev inequalities}
\label{sec:lsi}
We now use convexity in the head to obtain logarithmic Sobolev inequalities (LSIs). For fixed hidden weights, the logits depend linearly on the head $A$, so cross-entropy is convex in $A$. Exploiting this structure gives a global LSI for two-layer networks and, at greater depth, an LSI for the posterior conditioned on a spectral domain.

Throughout this section, constants are independent of $n$ and $m$.
They may depend on the model quantities in Table~\ref{tab:pi_lsi_constants}
and the input-norm bound $\bar{x}$ from Assumption~\ref{ass:inputs}.



\subsection{Two-layer networks: global LSI}
\label{sec:shallow_lsi}

\begin{theorem}[Shallow global LSI]
\label{thm:shallow_lsi}
Suppose Assumptions~\ref{ass:activation}--\ref{ass:inputs} hold and $L=2$. If $m/n^2$ is sufficiently large in terms of $C,\sigma,c_\phi,\bar{x},a_0$, the full posterior $\pi$ in~\eqref{eq:posterior} satisfies
\begin{equation}
\label{eq:shallow_lsi}
\CLS(\pi)\le\sigma^2+O\!\left(\frac n{\sqrt m}+\frac{n^2}{m}\right).
\end{equation}
Appendix~\ref{app:lsi_statements} gives an explicit finite-width bound.
\end{theorem}

\begin{rem}
When $m \gg n^2$, $C_{\mathrm{LS}}(\pi)$ converges to the Gaussian prior constant $\sigma^2$. The upper bound follows from the theorem; the matching lower bound is proved in Appendix~\ref{app:lsi_assembly}.
\qed
\end{rem}

\paragraph{Proof sketch.}
We bound the LSI constants of the conditional head and the hidden-weight
marginal, then combine them using a bound on how the conditional head law
changes with the hidden weights.

\paragraph{Step 1. Conditional head.}
Write $W=W_1$, let $\bar\pi$ be the marginal posterior of $W$, and let
$\mu_W$ be the conditional posterior of $A$ given $W$. Thus
\begin{equation}
\label{eq:lsi_disintegration}
\pi(dA,dW)=\mu_W(dA)\,\bar\pi(dW).
\end{equation}
For fixed $W$, cross-entropy is convex in $A$, so the conditional head potential has Hessian bounded below by $\sigma^{-2}I$. The Bakry--\'Emery criterion converts a uniform positive curvature bound into an LSI constant at most its reciprocal \citep[Section~5.7]{bakry2014analysis}, giving $\CLS(\mu_W)\le\sigma^2$.

\paragraph{Step 2. Hidden-weight marginal.}
Let $\bar U$ be the negative log-density of $\bar\pi$, and let
$\partial_v$ denote differentiation along a unit hidden-weight
direction $v$. Differentiating the integral over $A$ gives
\begin{equation}
\label{eq:marginal_curvature}
\begin{aligned}
\partial_v^2\bar U
    ={}&\sigma^{-2}+\E_{\mu_W}[\partial_v^2V]\\
       &-\Var_{\mu_W}(\partial_v V).
\end{aligned}
\end{equation}
The expectation contains nonnegative cross-entropy curvature and a feature remainder that can have either sign; the variance is the curvature lost as the conditional head distribution changes with $W$. The Brascamp--Lieb inequality \citep[Theorem~4.1]{brascamp1976extensions} bounds this variance by head derivatives weighted by the inverse conditional Hessian \citep[Eq.~(1.7)]{corderoerausquin2017transport}. The positive loss-curvature term and the variance bound both involve the change in logits along $v$, so we estimate them together. We retain the likelihood contribution to the conditional Hessian rather than bounding its inverse by $\sigma^2I$. Completing the square then combines the matching loss-curvature terms into a nonnegative quadratic form, leaving an interaction cost of order $\frac{n^2}{m}$. Thus only this cost and the nonlinear-feature remainder can reduce the marginal curvature. Appendix~\ref{app:lsi_curvature} gives the square-completion calculation.

The feature remainder contains products of a neuron's output weights with the loss gradient. At two layers, each hidden feature depends only on that neuron's incoming weights. Gaussian integration by parts in the corresponding head column, applied against the prior with the likelihood factor retained, transfers the head weight onto derivatives of the loss gradient and likelihood factor. Keeping the product inside its conditional expectation separates the bounded prior-mean contribution from these differentiated terms, each of which gains a factor $m^{-1/2}$. Summing the feature derivatives using Assumptions~\ref{ass:activation} and~\ref{ass:inputs}, and including the output normalization, bounds these contributions by $O(n/\sqrt m)$ and $O(n^2/m)$, respectively. 
Together with the interaction cost, this gives a marginal curvature loss
$\delta_m=O(n/\sqrt m+n^2/m)$. The integration-by-parts estimate is
proved in Appendix~\ref{app:lsi_curvature}. For the widths in the theorem,
$\delta_m<\sigma^{-2}$, and the Bakry--\'Emery criterion gives
$\CLS(\bar\pi)\le(\sigma^{-2}-\delta_m)^{-1}$.

\paragraph{Step 3. Assemble the joint LSI.}
To combine the two inequalities, decompose the entropy of a smooth
test function $g=g(A,W)$ as
\begin{equation}
\label{eq:lsi_entropy_decomposition}
\begin{aligned}
\Ent_\pi(g^2)
    ={}&\E_{\bar\pi}[\Ent_{\mu_W}(g^2)]\\
       &+\Ent_{\bar\pi}\!\left(\E_{\mu_W}[g^2]\right).
\end{aligned}
\end{equation}
The conditional LSI controls the first term. Applying the marginal LSI to the second requires differentiating the square root of the conditional second moment. Since $\mu_W$ also varies with $W$, its conditional score, the derivative of its log-density along $v$, contributes an additional term:
\[
\begin{aligned}
s_v(A)&=\E_{\mu_W}[\partial_vV]-\partial_vV(A,W),\\
\partial_v\E_{\mu_W}[g^2]
    &={}2\E_{\mu_W}[g\,\partial_v g]\\
    &\quad+\E_{\mu_W}[g^2s_v].
\end{aligned}
\]
These entropy and differentiation identities are recorded in \citet[Propositions~5.1--5.2]{srinivasan2026twoscale}. We control the score term by proving the following exponential-moment bound, uniformly in $W$ and unit directions $v$:
\begin{equation}
\label{eq:lsi_score_bound}
\begin{aligned}
\log\E_{\mu_W}e^{t s_v}&\le\frac{qt^2}{2},\qquad t\in\R,\\
q&=O\!\left(\frac{n^2}{m}\right).
\end{aligned}
\end{equation}
To prove~\eqref{eq:lsi_score_bound}, we compare $\mu_W$ with another head distribution. A transport inequality bounds the cost of moving between these laws by their relative entropy \citep[Proposition~1.1]{corderoerausquin2017transport}. The score difference splits into a term caused by head displacement and a term caused by the change in the loss gradient. The transport cost retains both Gaussian displacement of $A$ and the convex loss's excess above its tangent plane. The Gaussian part controls the first term, while the loss excess controls the second through the bounded Hessian of cross-entropy. Retaining both parts avoids requiring a uniform Lipschitz bound on the score as a function of $A$. Strong convexity bounds the conditional covariance of $A$, while bounded features and the Gaussian prior bound its conditional mean shift by $O(\sqrt n)$. Together, these moment estimates and transport costs bound the change in the expected score; the variational formula for entropy converts this into~\eqref{eq:lsi_score_bound}.

Entropy duality now controls the score term using the conditional entropy of $g^2$, which the conditional LSI bounds through derivatives with respect to $A$. Combining this with the conditional-moment differentiation identity and Cauchy--Schwarz gives
\[
\begin{aligned}
\norm{\nabla_W\bigl(\E_{\mu_W}[g^2]\bigr)^{1/2}}
\le{}&\bigl(\E_{\mu_W}[\norm{\nabla_Wg}^2]\bigr)^{1/2}\\
&+\sigma\sqrt q\,\bigl(\E_{\mu_W}[\norm{\nabla_Ag}^2]\bigr)^{1/2}.
\end{aligned}
\]
The second term accounts for the dependence of the conditional head law on $W$. Squaring this estimate when applying the marginal LSI in~\eqref{eq:lsi_entropy_decomposition} gives a correction of order $\sqrt q+q$. Combining the conditional and marginal inequalities through the two-scale framework of \citet[Theorem~3.1(3)]{srinivasan2026twoscale} therefore gives~\eqref{eq:shallow_lsi}. The curvature, score, and entropy calculations are given in Appendices~\ref{app:lsi_curvature}, \ref{app:lsi_score}, and~\ref{app:lsi_assembly}.

\qed

\begin{rem}[Input geometry]
\label{rem:lsi_input_geometry}
At fixed input norms, spreading the inputs across directions can reduce the width required by the two-layer LSI bound. For orthonormal inputs, $\CLS(\pi)\to\sigma^2$ whenever $m\gg n$. The proof is given in Appendix~\ref{app:lsi_input_geometry}.
\qed
\end{rem}

\subsection{Deep networks: local LSI on a region of posterior concentration}
\label{sec:deep_lsi}

At greater depth, hidden-weight perturbations propagate through subsequent matrices. We control this propagation using the same type of hidden spectral restriction as in~\eqref{eq:pi_cylinder}. Here we denote the scalar cutoff by $a$; it can be chosen independently of the global PI cutoff $\alpha$ and requires only $a>r_0+2\sigma$. Fix $L>2$ and such an $a$, independently of $n,m$, and define
\begin{equation}
\label{eq:lsi_domain}
\begin{aligned}
\mathcal G_a
    &=\bigcap_{\ell=2}^{L-1}
      \{\theta:\norm{W_\ell/\sqrt m}_{\op}\le a\},\\
\nu_a&=\pi(\cdot\mid\mathcal G_a).
\end{aligned}
\end{equation}
In this subsection, constants may also depend on the fixed depth $L$ and spectral cutoff $a$.
This domain is convex and leaves the head $A$ and first hidden matrix $W_1$ unrestricted. We first bound its posterior mass, before estimating the LSI constant.

\begin{lemma}[Posterior mass of the spectral domain]
\label{lem:spectral_mass}
Under Assumptions~\ref{ass:activation}--\ref{ass:prior_means}, fix $L>2$ and $a>r_0+2\sigma$. Then
\begin{equation}
\label{eq:spectral_mass}
\pi(\mathcal G_a^c)\le(L-2)e^{c_0n-c_1m},
\end{equation}
where $c_0,c_1>0$ are independent of $n,m$: $c_0$ depends only on $C,\sigma,a_0$, and $c_1$ only on $a,r_0,\sigma$.
\end{lemma}

\paragraph{Proof sketch.}
Bounded features and the Gaussian prior on $A$ give $\E_\gamma V\le c_0n$, so Jensen's inequality yields $Z\ge e^{-c_0n}$. Since $V\ge0$, this implies $\pi(E)\le e^{c_0n}\gamma(E)$ for every measurable event $E$. Gaussian operator-norm tails for the later hidden matrices, together with the prior-mean bound and a union bound, give~\eqref{eq:spectral_mass} with $c_1=(a-r_0-2\sigma)^2/(2\sigma^2)$ \citep[Theorem~5.32 and Proposition~5.34]{vershynin2012introduction}. The explicit choice of $c_0$ and full proof are recorded in Appendix~\ref{app:spectral_mass}.

\qed

\begin{theorem}[Deep local LSI]
\label{thm:deep_lsi}
Suppose Assumptions~\ref{ass:activation}--\ref{ass:inputs} hold. Fix depth $L>2$ and cutoff $a>r_0+2\sigma$. If $m/n^3$ is sufficiently large in terms of $C,\sigma,c_\phi,\bar{x},a_0,r_0,L,a$, the conditioned posterior $\nu_a$ in~\eqref{eq:lsi_domain} satisfies
\begin{equation}
\label{eq:deep_lsi}
\CLS(\nu_a)\le\sigma^2+
O_{L,a}\!\left(\frac{n^{3/2}}{\sqrt m}+\frac{n^2}{m}\right).
\end{equation}
The inequality uses the ambient parameter gradient restricted to
$\mathcal G_a$, and the first-layer center $W_{1,0}$ is arbitrary.
An explicit finite-width bound is given in Appendix~\ref{app:lsi_statements}.
\end{theorem}

\begin{rem}
When $m\gg n^3$, the theorem and the matching lower comparison in
Appendix~\ref{app:lsi_assembly} give $\CLS(\nu_a)\to\sigma^2$.
At fixed depth, the mass bound in Lemma~\ref{lem:spectral_mass} already
gives $\pi(\mathcal G_a)\to1$ whenever $c_1m-c_0n\to\infty$. Extending
the LSI to the full posterior requires additional entropy control
outside $\mathcal G_a$, which the PI gluing argument in
Remark~\ref{rem:pi_lsi_globalization} does not provide.
\qed
\end{rem}

\paragraph{Proof sketch.}
We use the same three steps as in the shallow case.

\paragraph{Step 1. Conditional head.}
Write $W=(W_1,\ldots,W_{L-1})$. For hidden weights in the spectral
domain, conditioning on $\mathcal G_a$ leaves the head law unchanged
because $\mathcal G_a$ restricts only hidden weights. Thus
$\mu_W=\nu_a(dA\mid W)=\pi(dA\mid W)$, and the conditional head
argument from Section~\ref{sec:shallow_lsi} gives $\CLS(\mu_W)\le\sigma^2$.

\paragraph{Step 2. Hidden-weight marginal.}
The marginal-curvature identity~\eqref{eq:marginal_curvature} applies
to the hidden marginal of $\nu_a$ in the interior of its domain.
The spectral bounds control propagation through the later layers,
giving feature-derivative estimates uniform in width at fixed $L,a$.

The nonlinear-feature remainder contains products of the head $A$ with second derivatives of the hidden features. Here we use projected head moments in place of the neuronwise integration-by-parts estimate. Appendix Lemma~\ref{lem:app_lsi_head} gives $\E_{\mu_W}\norm{Az}^2=O(n\norm z^2)$ for every fixed $z\in\R^m$. We take $z$ to be a second directional derivative of the hidden features, which is fixed when conditioning on $W$. The root-mean-square size of $Az$ is thus $O(\sqrt n)\norm z$; combining this with the feature-derivative bounds, the sum over $n$ observations, and the output normalization $m^{-1/2}$ gives a curvature loss of order $\frac{n^{3/2}}{\sqrt m}$. The remaining interaction cost is $O_{L,a}(\frac{n^2}{m})$.

The total curvature loss is
$\delta_m=O_{L,a}(n^{3/2}/\sqrt m+n^2/m)$. For the widths in the theorem,
the convex-domain Bakry--\'Emery criterion
\citep[Theorem~A1]{schlichting2019mixtures} gives a marginal LSI with
constant at most $(\sigma^{-2}-\delta_m)^{-1}$. The derivative estimates
are in Appendix~\ref{app:feature_derivatives}, and the domain
approximation is in Appendix~\ref{app:lsi_assembly}.

\paragraph{Step 3. Assemble the joint LSI.}
The score bound~\eqref{eq:lsi_score_bound} holds with
$q=O_{L,a}(n^2/m)$. Applying the entropy decomposition
\eqref{eq:lsi_entropy_decomposition} to $\nu_a$ and combining the
conditional and marginal LSIs yields~\eqref{eq:deep_lsi}. The term
$n^{3/2}/\sqrt m$ in the marginal-curvature loss gives the sufficient
width scale $m\gg n^3$. The score bound has the same order as in the
shallow case.

\qed

\begin{rem}[Depth and the prior scale]
\label{rem:depth_prior}
The theorem fixes $L$, but the derivative estimates quantify how depth enters the width requirement. A cutoff $a=1+\varepsilon$, with $\varepsilon>0$, is compatible with the mass bound when $r_0+2\sigma<1+\varepsilon$. For example, centered later-layer priors ($r_0=0$) with $\sigma=1/2$ allow every such cutoff. The leading second-derivative factor is then of order $L^2e^{2\varepsilon L}$, contributing a term of order $L^4e^{4\varepsilon L}n^3$ to the sufficient width. When $\varepsilon L$ is bounded, this contribution reduces to order $L^4n^3$. In the centered example, however, the mass exponent is $c_1=2\varepsilon^2$, so decreasing $\varepsilon$ increases the width needed for a prescribed posterior mass by a factor of order $\varepsilon^{-2}$. Appendix~\ref{app:lsi_depth} gives the full depth-dependent width condition.
\qed
\end{rem}




\section{Experiments}
\label{sec:experiments}
\begin{figure*}[!t]
\centering
\includegraphics[width=\textwidth]{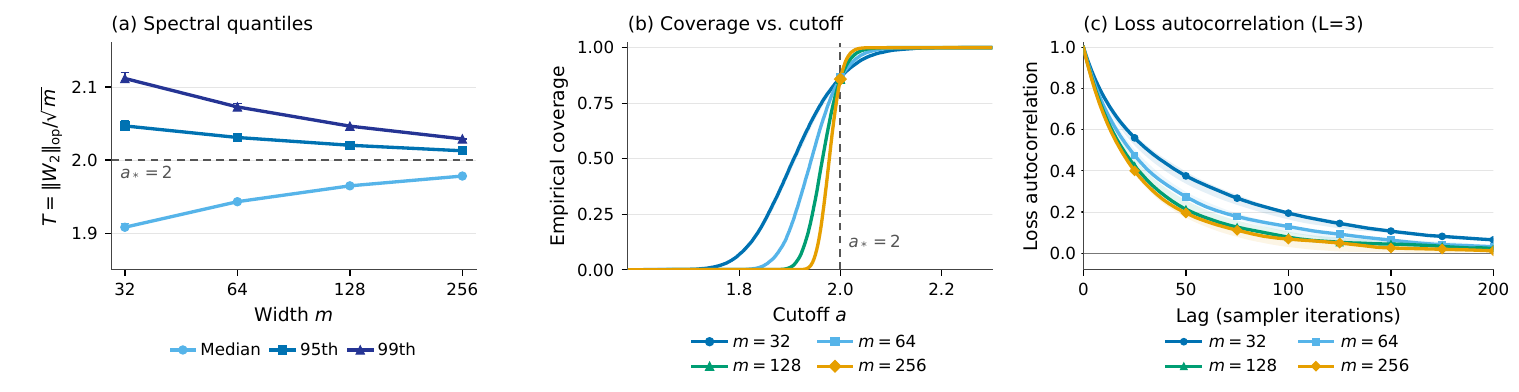}
\caption{\textbf{Spectral concentration and observable decorrelation with width.} All panels use networks with two hidden layers ($L=3$). (a) Empirical median, 95th, and 99th percentiles of $T=\norm{W_2}_{\op}/\sqrt m$. (b) Estimated posterior coverage of $\mathcal G_a=\{T\le a\}$ from elliptical slice samples. The dashed lines in (a,b) mark $r_0+2\sigma=2$; the deep LSI uses fixed cutoffs $a>2$. Table~\ref{tab:experimental_coverage} reports coverage at selected cutoffs. (c) Autocorrelation of summed training cross-entropy under the Metropolis-adjusted Gaussian-reference Langevin sampler at step size $0.01$, initialized from the reference chains. Both samplers target the unrestricted posterior. Points and curves show medians across three replicates. Autocorrelations are computed separately for four chains and averaged within replicate. Appendix~\ref{app:experiments_setup} gives the model, sampling, and estimator details.}
\label{fig:experiments}
\end{figure*}

We examine the mass captured by the deep LSI domain and the decorrelation of posterior observables as width increases. We consider binary classification with bias-free $\tanh$ networks having one or two hidden layers ($L=2$ or $3$), with the normalization in~\eqref{eq:network}. The data comprise $n=128$ unit-normalized training inputs, and the Gaussian prior has coordinate variance $\sigma^2=1$. Together with the chosen prior means, these settings satisfy Assumptions~\ref{ass:activation}--\ref{ass:inputs}. Within each of three replicates, all widths and both architectures use the same data and first-layer prior-center bank. Appendix~\ref{app:experiments_setup} gives the full setup and estimators.

\paragraph{Coverage of the spectral domain.}
The deep LSI in Theorem~\ref{thm:deep_lsi} holds for the posterior conditioned on the spectral domain in~\eqref{eq:lsi_domain}, so we first ask how much posterior mass this domain retains at the tested widths. For $L=3$, it is $\mathcal G_a=\{T\le a\}$, where $T=\norm{W_2}_{\op}/\sqrt m$. This controls propagation through the second hidden layer while leaving the head $A$ and first layer unrestricted. Since $r_0=0$ and $\sigma=1$, the admissible fixed cutoffs for the deep LSI satisfy $a>r_0+2\sigma=2$.

We estimate this coverage using elliptical slice samples of the full posterior \citep{murray2010elliptical}. Figure~\ref{fig:experiments}(a) shows the upper spectral quantiles moving toward the threshold as width grows. To see what this means for the mass retained by the domain, fix $a=2.05$. The corresponding coverage in Figure~\ref{fig:experiments}(b) rises from a median of $95.37\%$ at $m=32$ to $99.92\%$ at $m=256$; Table~\ref{tab:experimental_coverage} in Appendix~\ref{app:experimental_coverage} gives the full comparison. Because the comparison holds $a$ fixed, the growing coverage comes with the same control of feature propagation used in the local LSI argument. Increasing width therefore brings more posterior mass within an unchanged spectral restriction, in qualitative agreement with Lemma~\ref{lem:spectral_mass}.

\begin{rem}[Spectral scale and predictive information]
\label{rem:experimental_prediction}
The Gaussian prior and posterior have close spectral quantiles of $T$ (Table~\ref{tab:prior_spectral_comparison}). Posterior predictions improve the held-out log score over prior predictions at every tested width, with improvements of similar magnitude across widths within each architecture (Table~\ref{tab:predictive_scores}). These observations show that retaining a prior-like spectral scale is compatible with useful data-dependent changes in prediction.

\qed

\end{rem}

\paragraph{Observable decorrelation.}
To examine sampling behavior, we use the Gaussian-reference Langevin proposal of \citet[Section~4.3]{cotter2013mcmc} with a Metropolis--Hastings correction \citep{hastings1970monte}. We initialize four chains from the final states of the elliptical slice chains and use the same step size $0.01$ across widths. Tracking the summed training cross-entropy then gives a comparison of loss decorrelation per sampler iteration under the unrestricted posterior.

Figure~\ref{fig:experiments}(c) shows loss autocorrelation decreasing substantially with width for $L=3$. Wider parameterizations therefore exhibit less persistent loss fluctuations even as their parameter dimension grows. Held-out prediction probabilities show the same ordering between widths $m=32$ and $256$ (Figure~\ref{fig:prediction_acf}), extending the observation from the training objective to quantities used for inference. The loss ordering also survives step halving (Figure~\ref{fig:step_size}), showing stability under the tested change in numerical resolution.

The autocorrelation curves measure finite-lag relaxation of selected observables under the discrete sampler targeting $\pi$. Together with the domain-coverage estimates, they complement the entropy bound for $\nu_a$ in Theorem~\ref{thm:deep_lsi}, without directly estimating the LSI constant.

\bibliographystyle{plainnat}
\bibliography{bib}

\clearpage

\appendix

\section{Appendix: Global Poincar\'e bounds}
\label{app:PI}
\label{app:cylinder-b}

The proof of the global PI proceeds in three stages. We first establish the conditional bound in Theorem~\ref{thm:local}, using derivative estimates and exponential moments of the curvature deficit on a convex cylinder. Theorem~\ref{thm:global} removes the restrictions using Gaussian drift, and Section~\ref{app:global_pi_depth} tracks the depth dependence required for Theorem~\ref{thm:global_pi}. The remaining subsections give supplementary conditional inequalities and an alternative global two-layer LSI based on clipping. The sharper LSI bounds approaching $\sigma^2$ are proved in Appendix~\ref{app:lsi}.

\subsection{Model and conventions}
We use the model and data from Section~\ref{sec:model}, and write $k=C-1$ and $h=L-1$. In this notation, the network in~\eqref{eq:network} is
\begin{equation}
\label{eq:cylb-network}
\begin{aligned}
 H_1(x)&=\phi(W_1x),\\
 H_\ell(x)&=\phi(W_\ell H_{\ell-1}(x)/\sqrt m),\quad 2\le\ell\le h,\\
 f_\theta(x)&=AH_h(x)/\sqrt m.
\end{aligned}
\end{equation}
Parameter norms are the Euclidean product of Frobenius norms; $\norm{\cdot}_{\op}$ denotes the operator norm and $\norm{\cdot}_{\HS}$ the Hilbert--Schmidt norm.

\begin{assumption}[Activation and prior means]
\label{ass:cylb-model}
The activation $\phi\in C^\infty(\R)$ satisfies $\norm\phi_\infty\le1$, $\norm{\phi'}_\infty\le1$, and $\norm{\phi''}_\infty\le c_\phi<\infty$. The prior mean $\theta_0=(A_0,W_{1,0},\ldots,W_{h,0})$ is deterministic. There are finite bounds $a_0,r_0$, independent of width, such that
\[
 \norm{A_0}_{\op}\le a_0,\qquad
 \norm{W_{\ell,0}}_{\op}/\sqrt m\le r_0\quad(2\le\ell\le h).
\]
The first-layer mean $W_{1,0}$ is arbitrary. For $L=2$, no hidden-layer mean bound is required. When depth varies, the same bounds are used for all depths under consideration.
\end{assumption}
Throughout this appendix we use Assumption~\ref{ass:cylb-model}. For $\phi=\tanh$, one can take $c_\phi=4/(3\sqrt3)$. Write $a_j=Ae_j$, $a_{j,0}=A_0e_j$, and $b_0=\max_j\norm{a_{j,0}}\le a_0$.

The logits are $Qf_\theta(x_i)$, and the individual cross-entropy is
\[
 \ell_i(z)=\log\sum_{c=1}^C e^{(Qz)_c}-(Qz)_{y_i}.
\]
It satisfies $\ell_i\ge0$, $\norm{\nabla\ell_i}\le\sqrt2$, $\nabla^2\ell_i\succeq0$, and $\ell_i(z)\le\log C+2\norm z$. Write $f_i(\theta)=f_\theta(x_i)$, $H_{\ell,i}=H_\ell(x_i)$, $z_{1,i}=W_1x_i$, and $z_{\ell,i}=W_\ell H_{\ell-1,i}/\sqrt m$ for $\ell\ge2$. Set $X=\max\{1,\max_i\norm{x_i}\}$.

For prior variance $\sigma^2>0$ and a fixed loss multiplier $\tau\ge0$, set
\begin{equation}
\label{eq:cylb-posterior}
\begin{aligned}
 \gamma&=N(\theta_0,\sigma^2I),\qquad
 \V(\theta)=\tau\sum_{i=1}^n\ell_i(f_i(\theta)),\\
 Z_\gamma&=\E_\gamma e^{-\V},\qquad
 d\pi=Z_\gamma^{-1}e^{-\V}\,d\gamma,\\
 U(\theta)&=\norm{\theta-\theta_0}^2/(2\sigma^2)+\V(\theta).
\end{aligned}
\end{equation}
The fixed multiplier $\tau$ allows a rescaled likelihood, with $\tau=1$ recovering the posterior in~\eqref{eq:posterior}. We use the optimal constants $\CP$ and $\CLS$ from Definitions~\ref{def:pi} and~\ref{def:lsi}, and abbreviate integration by $\mu(q)=\int q\,d\mu$. Write $\lambda_P(\mu)=1/\CP(\mu)$ for the spectral gap. The Gaussian prior has $\CP(\gamma)=\CLS(\gamma)=\sigma^2$. The posterior generator is $\calL=\Delta-\nabla U\cdot\nabla$. Conditional inequalities use the Euclidean Dirichlet form with Neumann boundary conditions. Unless stated otherwise, data, output dimension, $\tau$, $\sigma$, and depth are fixed.

\subsection{Domains and posterior mass}
For head radius $B>0$ and scalar spectral threshold $\alpha>0$, we write the domains in~\eqref{eq:pi_cylinder} as
\begin{equation}
\label{eq:cylb-domains}
\begin{aligned}
 H_B&=\{\theta:\max_j\norm{a_j-a_{j,0}}\le B\},\\
 G_{B,\alpha}&=H_B\cap\bigcap_{\ell=2}^h
 \{\theta:\norm{W_\ell/\sqrt m}_{\op}\le \alpha\}.
\end{aligned}
\end{equation}
These domains are convex and leave $W_1$ unrestricted. Put $b=B+b_0$, so that $\max_j\norm{a_j}\le b$ on $H_B$. For $L=2$, the hidden intersection is empty and $G_{B,\alpha}=H_B$. For any measurable $G$ with $\pi(G)>0$, write $\pi_G=\pi(\cdot\mid G)$; in particular, $\nu=\pi(\cdot\mid G_{B,\alpha})$.

The following domination estimate transfers Gaussian tail and moment bounds to the posterior. This will control both the mass of the cylinder and the curvature moments used inside it. The near-Gaussian PI will require the separate curvature argument below.

\begin{lemma}[Domination and moments]\label{lem:mass}
Set
\[
 A_n=\tau n\bigl(\log C+2\sqrt{a_0^2+k\sigma^2}\bigr).
\]
Then $\pi\le e^{A_n}\gamma$. For each fixed $t>0$, $\gamma(e^{t\V})$ is bounded uniformly in width and depth when $a_0$ is bounded.
\end{lemma}
\begin{proof}
Conditional on all hidden weights, $f_i$ is Gaussian, with mean norm at most $a_0$ and covariance at most $\sigma^2I_k$, because $\norm{H_{h,i}}\le\sqrt m$. Hence $\gamma\V\le A_n$, and Jensen gives $Z_\gamma\ge e^{-A_n}$. For the second claim, use $\V\le\tau n\log C+2\tau\sum_i\norm{f_i}$ and H\"older across the $n$ observations. Each resulting Gaussian norm exponential moment is uniformly bounded.
\end{proof}
For $t\ge0$ and $B\ge\sigma(\sqrt k+\sqrt{2t})$, Gaussian concentration gives
$$
\pi(H_B^c)\le e^{A_n}m e^{-t}.
$$
If also $u\ge0$ and $\alpha\ge r_0+2\sigma+\sigma\sqrt{2u/m}$, a union bound gives
$$
\pi(G_{B,\alpha}^c)\le e^{A_n}\{m e^{-t}+(h-1)e^{-u}\}.
$$
Thus $B=O(\sqrt{\log(m/\delta)})$ and an appropriate spectral margin give mass at least $1-\delta$. A fixed $\alpha>r_0+2\sigma$ gives exponentially small hidden-layer exclusion for fixed depth.

\subsection{First and second derivatives}
The first derivative controls the entropy-transfer cost, while the second derivative identifies the potentially negative part of the posterior curvature. We keep the head-dependent backward columns explicit so that their Gaussian moments can be used later. Write $M_\ell=W_\ell/\sqrt m$, $D_{\ell,i}=\operatorname{Diag}\phi'(z_{\ell,i})$, and
\[
 T_{\ell,i}=AD_{h,i}M_h\cdots D_{\ell+1,i}M_{\ell+1},\qquad T_{h,i}=A.
\]
For fixed $\alpha>0$, define
\[
\begin{aligned}
 s_\ell&=X\left(\sum_{r=0}^{\ell-1}\alpha^{2r}\right)^{1/2},\\
 S_1&=\sum_{\ell=2}^h \alpha^{h-\ell}s_{\ell-1},\\
 S_2&=\sum_{\ell=1}^h \alpha^{h-\ell}s_\ell^2.
\end{aligned}
\]
An empty sum is zero. The quantity $s_\ell$ bounds first variations through layer $\ell$. The sums $S_1$ and $S_2$ collect, respectively, the mixed weight--feature terms and the terms containing $\phi^{\prime\prime}$ in the second derivative.

\begin{lemma}[Gradient estimate]\label{lem:gradient}
On $G_{B,\alpha}$,
\[
\begin{aligned}
 \norm{Df_i}_{\op}^2&\le1+b^2X^2\sum_{r=0}^{h-1}\alpha^{2r},\\
 \norm{\nabla\V}^2&\le M^2:=2\tau^2n^2
 \left(1+b^2X^2\sum_{r=0}^{h-1}\alpha^{2r}\right).
\end{aligned}
\]
\end{lemma}
\begin{proof}
For a direction $u$, $\norm{Dz_{1,i}[u]}\le X\norm{u_1}_F$ and $\norm{Dz_{\ell,i}[u]}\le\norm{u_\ell}_F+\alpha\norm{DH_{\ell-1,i}[u]}$. Cauchy--Schwarz over parameter blocks gives $\norm{Dz_{\ell,i}[u]},\norm{DH_{\ell,i}[u]}\le s_\ell\norm u$. Separately, $\norm{D_Af_i}_{\op}\le1$, $\norm{D_{W_1}f_i}_{\op}\le bX\alpha^{h-1}$, and $\norm{D_{W_\ell}f_i}_{\op}\le b\alpha^{h-\ell}$ for $\ell\ge2$, using $\norm A_{\op}\le b\sqrt m$. Square and sum these block bounds. Contract with $\tau\sum_i\nabla\ell_i$ to obtain the stated loss-gradient bound.
\end{proof}
For a vector-valued bilinear map $T$, write $\norm T_{\bil}=\sup_{\norm u=\norm v=1}\norm{T[u,v]}$.

The gradient estimate uses the deterministic head cutoff. For curvature, we retain the largest normalized backward column as the random quantity $Z_*$, whose exponential moments are estimated in Lemma~\ref{lem:tail}.

\begin{lemma}[Pointwise random curvature envelope]\label{lem:hessian}
Define on $G_{B,\alpha}$
\[
 Z_*(\theta)=\max_{i,\ell,j}\alpha^{-(h-\ell)}\norm{T_{\ell,i}e_j}.
\]
Then
\[
 \norm{D^2f_i}_{\rm bil}\le\frac{2s_h+2bS_1+c_\phi Z_*S_2}{\sqrt m}.
\]
Consequently, with $U=\norm{\theta-\theta_0}^2/(2\sigma^2)+\V$,
$$
\begin{aligned}
D(\theta)&:=\max\{0,\sigma^{-2}
    -\lambda_{\min}(\nabla^2U(\theta))\},\\
D(\theta)&\le\frac{d_0+d_1Z_*(\theta)}{\sqrt m},\\
\nabla^2U(\theta)&\succeq
    (\sigma^{-2}-D(\theta))I.
\end{aligned}
$$
Here,
\[
 d_0=\sqrt2\tau n(2s_h+2bS_1),\qquad
 d_1=\sqrt2\tau n c_\phi S_2.
\]
\end{lemma}
\begin{proof}
For unit directions $u,v$, set
$$
P_{\ell,i}=\phi''(z_{\ell,i})\odot Dz_{\ell,i}[u]\odot Dz_{\ell,i}[v].
$$
The exact second derivative is
\begin{align*}
 \sqrt m D^2f_i[u,v]
 ={}&u_A DH_{h,i}[v]+v_A DH_{h,i}[u]\\
 &+\sum_{\ell=1}^h T_{\ell,i}P_{\ell,i}\\
 &+m^{-1/2}\sum_{\ell=2}^h T_{\ell,i}D_{\ell,i}u_\ell DH_{\ell-1,i}[v]\\
 &+m^{-1/2}\sum_{\ell=2}^h T_{\ell,i}D_{\ell,i}v_\ell DH_{\ell-1,i}[u].
\end{align*}
For the nonlinear insertions use $\norm{T(p\odot q)}\le\max_j\norm{Te_j}\norm p\norm q$. For the mixed insertions use $\norm{T_{\ell,i}D_{\ell,i}}_{\op}\le b\sqrt m\,\alpha^{h-\ell}$. The remaining two terms are at most $2s_h$. This gives the predictor bound. The loss Hessian is
\[
 \nabla^2\V=\tau\sum_i\left[(Df_i)^*\nabla^2\ell_i Df_i+
 \langle\nabla\ell_i,D^2f_i\rangle\right].
\]
The first term is positive semidefinite and $\norm{\nabla\ell_i}\le\sqrt2$.
\end{proof}
Since sublevel sets of $Z_*$ need not be convex, we estimate its moments on $G_{B,\alpha}$ to retain the convex boundary condition.

\subsection{Conditional Poincar\'e inequality}
We first obtain a defective entropy inequality by changing density from the conditioned Gaussian prior. Combined with an exponential moment of the curvature deficit, it will control the weighted error in the Bochner identity.

\begin{lemma}[Defective entropy transfer]\label{lem:dlsi}
Let $\mu$ satisfy $\CLS(\mu)\le\sigma^2$ on a convex domain, $d\nu=Z^{-1}e^{-V}d\mu$, and $\norm{\nabla V}\le M$. Then
\[
 \Ent_\nu(g^2)\le8\sigma^2\nu(|\nabla g|^2)+3\sigma^2M^2\nu(g^2).
\]
\end{lemma}
\begin{proof}
Apply the reference LSI to $g e^{-V/2}/\sqrt Z$. The elementary square inequality yields
\[
\begin{aligned}
 \Ent_\nu(g^2)\le{}&4\sigma^2\nu(|\nabla g|^2)+\sigma^2M^2\nu(g^2)\\
 &+\nu[g^2(V+\log Z)].
\end{aligned}
\]
Entropy duality bounds the last term by
\[
 \tfrac12\Ent_\nu(g^2)+\tfrac12\log\{\mu(e^V)\mu(e^{-V})\}\nu(g^2).
\]
The Herbst consequence of the reference LSI gives the logarithm at most $\sigma^2M^2$. Absorb the half entropy. Convexity converts the gradient bound into a Lipschitz bound, so Herbst applies to $V$ on the domain.
\end{proof}

\begin{rem}[From curvature to a Poincar\'e inequality]
\label{rem:bochner_background}
To explain how curvature yields a PI, first consider a smooth bounded convex domain $\Omega$ with conditioned density $\nu_\Omega\propto e^{-U}$. Write $\calL g=\Delta g-\langle\nabla U,\nabla g\rangle$. For a smooth $g$ with zero outward normal derivative, the integrated Bochner identity \citep{bakry2014analysis,kolesnikov2017brascamp} gives
\begin{equation}
\label{eq:pi_bochner_sketch}
\begin{aligned}
\int(\calL g)^2\,d\nu_\Omega
={}&\int\norm{\nabla^2g}_{\HS}^2\,d\nu_\Omega\\
 &+\int\langle\nabla^2U\nabla g,\nabla g\rangle\,d\nu_\Omega
 +\mathcal B_\Omega(g),
\end{aligned}
\end{equation}
where $\norm{\cdot}_{\HS}$ is the Frobenius norm and $\mathcal B_\Omega(g)\ge0$ is a boundary term, nonnegative by convexity. The two interior terms measure the variation of the gradient and the curvature of the potential. On a Riemannian manifold there is also a Ricci-curvature term; it vanishes in our Euclidean parameter space. If $\nabla^2U\succeq\rho I$ everywhere for some $\rho>0$, this identity gives $\int(\calL g)^2\,d\nu_\Omega\ge\rho\int\norm{\nabla g}^2\,d\nu_\Omega$. For a nonconstant Neumann eigenfunction $-\calL g=\lambda g$, integration by parts gives $\int\norm{\nabla g}^2\,d\nu_\Omega=\lambda\int g^2\,d\nu_\Omega$. Thus $\lambda^2\ge\rho\lambda$, so the spectral gap is at least $\rho$ and $\CP(\nu_\Omega)\le\rho^{-1}$. Here, substituting~\eqref{eq:pi_curvature_deficit} instead produces the error $\int D\norm{\nabla g}^2\,d\nu_\Omega$.
\end{rem}

The next theorem makes this step independent of the network model. Its inputs are a defective entropy bound and exponential moments of $D$. Its conclusion is a spectral-gap bound, obtained by absorbing part of the weighted curvature error into the Hessian term of the Bochner identity.

\begin{theorem}[Curvature-defect PI transfer]\label{thm:transfer}
Let $\Omega$ be a bounded smooth convex domain and $U$ be smooth on its closure, so that $\nu\propto e^{-U}$ has positive density and a self-adjoint Neumann generator. Suppose
\[
 \Ent_\nu(g^2)\le A\nu(|\nabla g|^2)+b_D\nu(g^2),\qquad
 \nabla^2U\succeq(\rho-D)I,
\]
with $D\ge0$. Then
\[
 \lambda_P(\nu)\ge\rho-
 \inf_{t\ge A}\frac{b_D+\log\nu(e^{tD})}{t},
\]
whenever the right side is positive.
\end{theorem}
\begin{proof}
Entropy duality gives
\[
 \nu(Dg^2)\le\frac A t\nu(|\nabla g|^2)
 +\frac{b_D+\log\nu(e^{tD})}{t}\nu(g^2).
\]
For a smooth Neumann function $f$, use the integrated Bochner identity \eqref{eq:pi_bochner_sketch}; Remark~\ref{rem:bochner_background} explains its curvature terms and the spectral-gap argument. To make the boundary sign explicit, let $n$ be the outward unit normal, $\mathrm{II}(v,w)=\langle D_vn,w\rangle$ the second fundamental form, and $d\nu_{\partial\Omega}=Z_\Omega^{-1}e^{-U}\,dS$ the weighted surface measure, where $Z_\Omega=\int_\Omega e^{-U}\,d\theta$. The boundary term is $\int_{\partial\Omega}\mathrm{II}(\nabla_Tf,\nabla_Tf)\,d\nu_{\partial\Omega}$, with $\nabla_Tf$ the tangential gradient. It is nonnegative by convexity \citep[Theorem~1.1]{kolesnikov2017brascamp}. Substituting $\nabla^2U\succeq(\rho-D)I$ therefore gives
\[
 \nu((\calL f)^2)\ge\nu(\norm{\nabla^2f}_{\HS}^2)
 +\rho\nu(|\nabla f|^2)-\nu(D|\nabla f|^2).
\]
Use $g=|\nabla f|$, regularized at zero, and $|\nabla|\nabla f||^2\le\norm{\nabla^2f}_{\HS}^2$. For $t\ge A$ the Hessian term can be dropped. The spectral theorem gives the gap bound. Constants are the only zero-energy functions because the domain is connected and the density positive. The same argument applies on other domains whenever a Neumann operator core justifies the integrated Bochner inequality and its extension to the generator domain. The cylinder application is obtained through the convex approximation constructed in the proof of Theorem~\ref{thm:local}.
\end{proof}

To apply Theorem~\ref{thm:transfer}, it remains to estimate the curvature moments through the envelope in Lemma~\ref{lem:hessian}. Conditioning on the hidden weights makes each backward column a Gaussian projection of the head. The following bound combines this observation with posterior domination from Lemma~\ref{lem:mass}.

\begin{lemma}[Backward-column tails and their exponential moments]\label{lem:tail}
Let $\nu=\pi(\cdot\mid G_{B,\alpha})$, $p_G=\pi(G_{B,\alpha})>0$, and $N=nhm$. Put
\[
 b_*=a_0+\sigma\left(\sqrt k+
 \sqrt{2\log(Ne^{A_n}/p_G)}\right).
\]
Then
\[
\begin{aligned}
 \nu(Z_*>b_*+u)&\le e^{-u^2/(2\sigma^2)}\quad(u\ge0),\\
 \log\nu(e^{sZ_*})&\le sb_*+\sigma^2s^2+\log2\quad(s\ge0).
\end{aligned}
\]
\end{lemma}
\begin{proof}
Under the prior, condition on all hidden layers satisfying their spectral constraints. Each normalized backward column is $Av$ with $\norm v\le1$. Its mean norm is at most $a_0$ and its Gaussian covariance at most $\sigma^2I_k$. Apply Gaussian norm concentration, a union bound over at most $N$ columns, and $\nu(E)\le e^{A_n}\gamma(E\cap G_{B,\alpha})/p_G$. This gives the first assertion. For $Y=(Z_*-b_*)_+$, the tail implies
\[
 \nu(e^{Y^2/(4\sigma^2)})
 =1+\int_1^\infty\nu(Y>2\sigma\sqrt{\log u})\,du\le2.
\]
Since $sY\le Y^2/(4\sigma^2)+\sigma^2s^2$, the moment bound follows.
\end{proof}

We now combine the derivative, entropy, and moment estimates. The defect $\Delta_{m,L}$ below records their total cost relative to the Gaussian curvature. The proof also passes from smooth bounded domains to the original cylinder.

\begin{theorem}[Explicit conditional deep PI]\label{thm:local}
With the preceding definitions, put
\[
 \Delta_{m,L}=\frac{d_0+d_1b_*}{\sqrt m}
 +\frac{2\sigma d_1}{\sqrt m}\sqrt{3\sigma^2M^2+\log2}
 +\frac{8\sigma^4d_1^2}{m}.
\]
If $\Delta_{m,L}<\sigma^{-2}$, then
\[
 \lambda_P(\nu)\ge\sigma^{-2}-\Delta_{m,L},\qquad
 \CP(\nu)\le(\sigma^{-2}-\Delta_{m,L})^{-1}.
\]
Also
\[
 \CLS(\nu)\le4\sigma^2+
 \frac{3\sigma^2M^2+2}{2(\sigma^{-2}-\Delta_{m,L})}.
\]
\end{theorem}
\begin{proof}
The truncated Gaussian reference has LSI constant at most $\sigma^2$. Lemmas~\ref{lem:gradient} and \ref{lem:dlsi} give $A=8\sigma^2$ and $b_D=3\sigma^2M^2$. Lemma~\ref{lem:tail} gives
\[
 \log\nu(e^{tD})\le t(d_0+d_1b_*)/\sqrt m
 +\sigma^2d_1^2t^2/m+\log2.
\]
Apply Theorem~\ref{thm:transfer} and choose $t=8\sigma^2+\sqrt{m(3\sigma^2M^2+\log2)/(\sigma^2d_1^2)}$. The case $d_1=0$ follows by sending $t$ to infinity.

For the domain justification, fix $m$ and choose smooth bounded convex inner approximations $\Omega_j\subset G_{B,\alpha}^{\circ}$ whose indicators converge almost everywhere to $\mathbf1_{G_{B,\alpha}}$. Such an exhaustion is obtained from bounded convex inner bodies followed by smoothing of their boundaries. Set
\[
 p_j=\pi(\Omega_j),\qquad \nu_j=\pi(\cdot\mid\Omega_j),\qquad
 \mu_j=\gamma(\cdot\mid\Omega_j).
\]
Convex-domain Bakry--\'Emery theory \citep{bakry2014analysis} gives $\CLS(\mu_j)\le\sigma^2$. The pointwise derivative estimates hold on every $\Omega_j$ with the same $M,d_0,d_1$. Apply Lemma~\ref{lem:dlsi} separately to $(\mu_j,\nu_j)$ to obtain the defective LSI there. For the moment estimate,
\[
 \nu_j(E)\le\frac{e^{A_n}}{p_j}\gamma(E\cap\Omega_j).
\]
Since $\Omega_j\subset G_{B,\alpha}$, the backward-column Gaussian calculation applies after dropping the additional constraints. Lemma~\ref{lem:tail} therefore holds with $p_G$ replaced by $p_j$. As $p_j\to p_G>0$, the thresholds $b_{*,j}$ and resulting defects $\Delta_{m,L,j}$ converge to $b_*$ and $\Delta_{m,L}$. For all sufficiently large $j$ the gap bound is positive. Apply the smooth-domain PI to restrictions of ambient smooth compactly supported functions and pass to the limit by dominated convergence. Cutoffs and local Sobolev density on convex Lipschitz domains extend it to $H^1(G_{B,\alpha},\nu)$. The defective LSI follows directly from the conditioned Gaussian reference, or by lower semicontinuity along the same exhaustion. Rothaus' inequality gives the LSI bound.
\end{proof}
For fixed depth, $\alpha,a_0,r_0$, and $p_G\ge p_0>0$, this yields
\[
 \Delta_{m,L}=O_L\left(\frac{B+1+\sqrt{\log m}}{\sqrt m}+\frac1m\right).
\]
In particular, at $B\asymp\sqrt{\log m}$,
\[
\begin{aligned}
 \CP(\nu)&\le\sigma^2+O_L(\sqrt{\log m/m}),\\
 \CLS(\nu)&\le O_L(1+\log m).
\end{aligned}
\]

\subsection{Global Poincar\'e inequality}
We now remove the restrictions while preserving the conditional PI from Theorem~\ref{thm:local}. We release hidden layers in increasing order: upstream features remain bounded by the activation, while downstream spectral bounds remain available. This keeps the likelihood gradient controlled in the block being released, allowing Gaussian drift to dominate outside its cutoff. The argument uses the Lyapunov approach to Poincar\'e inequalities \citep{bakry2008lyapunov}.

\begin{lemma}[Gluing]\label{lem:glue}
On a connected domain $\Omega$, let $G\subset\Omega$ have a conditional PI. Suppose a positive function $W$ satisfies
\[
\begin{aligned}
 -\calL W/W&\ge\mathfrak a>0\quad\text{on }\Omega\setminus G,\\
 -\calL W/W&\ge-\mathfrak b\quad\text{on }\Omega.
\end{aligned}
\]
Assume the boundary condition permitting $\pi_\Omega((-\calL W/W)g^2)\le\pi_\Omega(|\nabla g|^2)$; outward $\partial_nW\ge0$ suffices. Then
\[
 \CP(\pi_\Omega)\le\mathfrak a^{-1}
 +(1+\mathfrak b/\mathfrak a)\CP(\pi_G).
\]
\end{lemma}
\begin{proof}
Subtract the mean on $G$. The drift controls the exterior square integral by $\mathfrak a^{-1}$ times the energy plus $\mathfrak b/\mathfrak a$ times the interior square integral. Apply the conditional PI to the latter. The unrestricted variance is no larger than this square integral.
\end{proof}

In Lemma~\ref{lem:glue}, the exterior drift determines the additive extension cost, while the ratio of the negative drift to the exterior drift determines the multiplicative cost. The next lemma gives an exterior drift of order $m$ and an exponentially smaller negative part for each hidden layer.

\begin{lemma}[Normalized hidden-layer Lyapunov function]\label{lem:net}
Suppose downstream layers remain bounded by $\alpha$, the head remains in $H_B$, and $2\le\ell\le h$. Then
\[
 \norm{\nabla_{W_\ell}\V}_F\le G_\ell:=\sqrt2\tau n b \alpha^{h-\ell}.
\]
Let $r_\ell=\norm{W_{\ell,0}}_{\op}/\sqrt m$ and assume $\alpha>r_\ell$. Fix $\varepsilon\in(0,1)$, $\eta\in(0,(2\sigma^2)^{-1})$, and assume
\[
 (1-\varepsilon)^2(\alpha-r_\ell)^2>
 \frac1{\sigma^{-2}-2\eta}+\frac{\log(1+2/\varepsilon)}\eta.
\]
When centers depend on width, assume that the difference between the left and right sides of this inequality has a positive lower bound independent of $m$. If $G_\ell^2=o(m)$, there is a positive Lyapunov function with exterior drift at least $cm$ outside the $\ell$th spectral ball and global negative drift at most $Cm e^{-c'm}$. Its normal derivative is zero on all remaining product boundaries.
\end{lemma}
\begin{proof}
The gradient bound uses only the bounded upstream activations and downstream spectral norms. Let $\mathcal N_\varepsilon$ be a sphere net of size at most $e^{md_\varepsilon}$, $d_\varepsilon=\log(1+2/\varepsilon)$, and $q_v=\norm{(W_\ell-W_{\ell,0})v}^2$. For $c_0=\sigma^{-2}-2\eta$,
\[
 -\calL e^{\eta q_v}/e^{\eta q_v}
 \ge2\eta(c_0q_v-m-G_\ell\sqrt{q_v}).
\]
Choose a small fixed $\zeta>0$, put $c_*=c_0-\zeta$, $m'=m+G_\ell^2/(4\zeta)$, and choose $s$ strictly between
\[
 c_*^{-1}+d_\varepsilon/\eta
 \quad\text{and}\quad A_*=(1-\varepsilon)^2(\alpha-r_\ell)^2.
\]
The uniform strict margin permits these choices with constants independent of width. Set
\[
 \widetilde F=1+e^{-\eta sm}\sum_{v\in\mathcal N_\varepsilon}e^{\eta q_v}.
\]
Since $G_\ell\sqrt q\le\zeta q+G_\ell^2/(4\zeta)$, negative summands occur only for $q_v<m'/c_*$. Thus
\[
 -\calL\widetilde F/\widetilde F
 \ge-2\eta m'e^{-\eta sm+md_\varepsilon+\eta m'/c_*}
 \ge-Cm e^{-c'm}.
\]
Outside the spectral ball, $\alpha>r_\ell$ gives $\norm{W_\ell-W_{\ell,0}}_{\op}>(\alpha-r_\ell)\sqrt m$, hence $\max_vq_v\ge A_*m$. With $S=\widetilde F-1$, the exponential-weight identity gives
\[
 \frac{\sum_v e^{\eta q_v}q_v}{\sum_v e^{\eta q_v}}
 \ge\max_vq_v-\frac{\log|\mathcal N_\varepsilon|}{\eta}.
\]
Therefore
\[
 -\calL\widetilde F/\widetilde F\ge
 2\eta\frac S{1+S}[c_*(A_*-d_\varepsilon/\eta)m-m']\ge cm,
\]
since $S\ge e^{\eta(A_*-s)m}$ and $m'/m\to1$. As $\widetilde F$ depends only on $W_\ell$, its normal derivative vanishes on the boundaries of the other parameter blocks. Integration by parts holds for compactly supported test functions; Gaussian tails and $\eta<(2\sigma^2)^{-1}$ justify passage to the limit along smooth cutoffs.
\end{proof}
Optimizing the strict net condition gives the sufficient numerical buffer
\[
\begin{aligned}
 \alpha-r_\ell>{}&\inf_{0<\varepsilon<1}
 \frac{\sigma[1+\sqrt{2\log(1+2/\varepsilon)}]}{1-\varepsilon}
 \\={}&3.852887\ldots\,\sigma.
\end{aligned}
\]
This buffer comes from the sphere-net Lyapunov construction. It is stronger than the $2\sigma$ margin used for posterior concentration in Lemma~\ref{lem:mass}, because restriction removal requires a drift estimate as well as a tail bound.

\begin{theorem}[Global PI at every fixed depth]\label{thm:global}
Under the standing assumptions of this appendix, fix $L\ge2$. For $L>2$, choose a fixed $\alpha>r_0+3.853\sigma$; at $L=2$ no hidden spectral cutoff is required. Then, for sufficiently large width,
\[
\begin{aligned}
 \CP(\pi)&\le\sigma^2+O_L(m^{-1/3}),\\
 \lambda_P(\pi)&\ge\sigma^{-2}-O_L(m^{-1/3}).
\end{aligned}
\]
\end{theorem}
\begin{proof}
Start with $G_{B,\alpha}$ and release $W_2,W_3,\ldots,W_h$ in this order. Downstream bounds remain valid at each step. If $B=o(\sqrt m)$, Lemma~\ref{lem:net} applies. Lemma~\ref{lem:glue} costs $O_L(1/m)$ per layer and a multiplier $1+O_L(e^{-cm})$. Thus
\[
 \CP(\pi_{H_B})\le\sigma^2+
 O_L((B+1+\sqrt{\log m})/\sqrt m+1/m).
\]
The required lower bound on $p_G$ follows from Lemma~\ref{lem:mass} for the choices below. For the head, $\norm{\nabla_{a_j}\V}\le h_m=\sqrt2\tau n/\sqrt m$ globally. Set
\[
 q_j=\norm{a_j-a_{j,0}}^2/(4\sigma^2),\qquad
 W=m^{-1}\sum_j e^{q_j}.
\]
Completion of the square gives
\[
 -\calL W/W\ge
 \frac{\sum_j e^{q_j}q_j}{2\sigma^2\sum_j e^{q_j}}
 -\frac{k+\sigma^2h_m^2}{2\sigma^2}.
\]
The weighted average is at least $\max_jq_j-\log m$. Set
\[
 B_\omega^2=4\sigma^2(\log m+k+\sigma^2h_m^2+\omega).
\]
The exterior drift is $\omega/(2\sigma^2)$ and the negative global drift is at most $(k+\sigma^2h_m^2)/(2\sigma^2)$. Gluing yields
\[
 \CP(\pi)\le\sigma^2+O_L\left(\omega^{-1}
 +\sqrt{(\omega+\log m)/m}+m^{-1}\right).
\]
Take $\omega=m^{1/3}$. Then $B_\omega=O(m^{1/6})$, all hidden-layer conditions hold, and all inverse-gap denominators are positive for large width.
\end{proof}

\subsection{Explicit depth dependence of the global PI}
\label{app:global_pi_depth}

We first record the depth dependence of the local derivative estimates. These estimates allow cutoffs beyond the regime used for global restriction removal.

\begin{lemma}[Dependence on depth]\label{lem:depth-dependence}
For fixed $\alpha<1$, the derivative sums are bounded in depth; for $\alpha=1$ they give $s_h=O(\sqrt L)$, $S_1=O(L^{3/2})$, $S_2=O(L^2)$; for fixed $\alpha>1$ they are bounded by constants times $\alpha^L,L\alpha^L,\alpha^{2L}$ respectively. For $\alpha>1$ and $h\ge2$,
\[
 X(h-1)\alpha^{h-2}\le S_1\le\frac{X(h-1)\alpha^{h-2}}{\sqrt{1-\alpha^{-2}}}.
\]
The corresponding errors in $\Delta_{m,L}$ are of order
\[
 O\left(L^{5/2}\sqrt{\log(mL)/m}+L^4/m\right)\quad(\alpha=1),
\]
\[
 O_\alpha\left(\alpha^{3L}\sqrt{\log(mL)/m}+\alpha^{4L}/m\right)\quad(\alpha>1),
\]
when the other quantities and the domination exponent are controlled and $B,b_*=O(\sqrt{\log(mL)})$. For $\alpha_m=1+t_m$, the $\alpha=1$ orders persist if $Lt_m=O(1)$.
\end{lemma}
\begin{proof}
For $\alpha<1$, bound the defining geometric sums by their infinite sums. For $\alpha=1$, $s_\ell=X\sqrt\ell$, so summing gives $S_1=O(L^{3/2})$ and $S_2=O(L^2)$. For $\alpha>1$, use $X\alpha^{\ell-1}\le s_\ell\le X\alpha^{\ell-1}/\sqrt{1-\alpha^{-2}}$ in the definitions of $S_1,S_2$. Substitution of these estimates and the corresponding bound on $M$ into $\Delta_{m,L}$ gives the stated errors. If $\alpha_m=1+t_m$ and $Lt_m=O(1)$, the factors $\alpha_m^\ell$ stay bounded for $\ell\le L$, giving the same orders as at $\alpha=1$.
\end{proof}

We specialize to $\tau=1$ and hold the dataset, $C,\sigma,c_\phi,a_0,r_0$ fixed as depth and width vary. Fix $\alpha>\max\{1,r_0+3.853\sigma\}$. The following theorem makes the local curvature bound, the drift conditions, and the cylinder-mass estimate hold under a single width threshold, with constants independent of depth. Constants in this subsection may depend on these fixed quantities, but not on $L,m$.

\begin{theorem}[Global PI with explicit depth dependence]
\label{thm:cylb-global-depth}
Under Assumption~\ref{ass:cylb-model}, with $\tau=1$ and the fixed quantities and cutoff specified above, there are constants $K,K_0>0$, independent of $L,m$, such that for every integer $L>2$,
\[
 m\ge K_0\alpha^{9L}\quad\Longrightarrow\quad
 \CP(\pi)\le\sigma^2+K \alpha^{3L}m^{-1/3}.
\]
The first-layer prior mean remains arbitrary.
\end{theorem}
\begin{proof}
Put $h=L-1$ and $j=L-2$. We use $K_1$ for a positive constant independent of $L,m$, increasing it when needed.

\paragraph{Local PI.}
The geometric sums defining $s_h,S_1,S_2$ give
$$
s_h\le K_1\alpha^h,\qquad
S_1\le K_1h \alpha^h,\qquad
S_2\le K_1\alpha^{2h}.
$$
Since $\alpha>1$, $h \alpha^h\le K_1\alpha^{3h}$. If $p_G=\pi(G_{B,\alpha})\ge1/2$, the definitions of $M,d_0,d_1,b_*$ imply
$$
\begin{aligned}
M&\le K_1(1+B)\alpha^h,\\
d_0&\le K_1\{\alpha^h+(1+B)h \alpha^h\},\\
d_1&\le K_1\alpha^{2h},\\
b_*&\le K_1\{1+\sqrt{\log(mL)}\}.
\end{aligned}
$$
Substituting these bounds in Theorem~\ref{thm:local} gives
\begin{equation}
\label{eq:global_depth_delta}
\begin{aligned}
\Delta_{m,L}\le K_1\Bigl\{
&\frac{(1+B)\alpha^{3L}}{\sqrt m}\\
&+\frac{\alpha^{2L}(1+\sqrt{\log(mL)})}{\sqrt m}
+\frac{\alpha^{4L}}m\Bigr\}.
\end{aligned}
\end{equation}
When $\Delta_{m,L}\le(2\sigma^2)^{-1}$, the local PI therefore gives
$$
\CP(\nu)\le\sigma^2+2\sigma^4\Delta_{m,L}.
$$

\paragraph{Removing the hidden restrictions.}
We choose the parameters in Lemma~\ref{lem:net} once, independently of $L,m$. The strict cutoff margin permits $\varepsilon\in(0,1)$ and $\eta\in(0,(2\sigma^2)^{-1})$ such that, with
$$
A_*=(1-\varepsilon)^2(\alpha-r_0)^2,
\qquad d=\log(1+2/\varepsilon),
$$
we have
$$
A_*>\frac1{\sigma^{-2}-2\eta}+\frac d\eta.
$$
Choose $\zeta>0$ so that $c_*:=\sigma^{-2}-2\eta-\zeta>0$ and $A_*>c_*^{-1}+d/\eta$. Next choose
$$
c_*^{-1}+d/\eta<s<A_*,
\qquad q_*:=\frac{c_*(s-d/\eta)-1}{2}>0.
$$
The proof of Lemma~\ref{lem:net} then gives constants
$$
\begin{aligned}
u&=\eta\{c_*(A_*-d/\eta)-1-q_*\}>0,\\
v&=2\eta(1+q_*)>0,\\
\kappa&=\eta\{s-d/\eta-(1+q_*)/c_*\}>0.
\end{aligned}
$$
Indeed, if $G_\ell^2/m\le4\zeta q_*$, its quantity $m'=m+G_\ell^2/(4\zeta)$ is at most $(1+q_*)m$. The same Lyapunov function has exterior drift at least $um$ and negative part at most $vm e^{-\kappa m}$. These bounds hold for every released layer, since its normalized prior mean is at most $r_0$. The gradient estimate gives
\begin{equation}
\label{eq:global_depth_gradient}
\max_{2\le\ell\le h}\frac{G_\ell^2}{m}
\le K_1\frac{(1+B)^2\alpha^{2L}}m.
\end{equation}
Put $e_m=(v/u)e^{-\kappa m}$. Iterating Lemma~\ref{lem:glue} over the $j$ hidden restrictions yields
$$
\begin{aligned}
\CP(\pi_{H_B})
&\le(1+e_m)^j\CP(\nu)
+\frac1{um}\sum_{r=0}^{j-1}(1+e_m)^r\\
&\le e^{j e_m}\left\{\CP(\nu)+\frac{j}{um}\right\}.
\end{aligned}
$$
Using the local PI and $e^x-1\le xe^x$ for $x\ge0$, we obtain
\begin{equation}
\label{eq:global_depth_hidden}
\CP(\pi_{H_B})\le\sigma^2+
e^{j e_m}\left\{2\sigma^4\Delta_{m,L}
+\frac{j}{um}+\sigma^2j e_m\right\}.
\end{equation}

\paragraph{Removing the head restriction.}
Choose the radius from the proof of Theorem~\ref{thm:global}:
$$
\begin{aligned}
\omega&=m^{1/3},\qquad h_m=\sqrt2n/\sqrt m,\\
B^2&=4\sigma^2\{\log m+k+\sigma^2h_m^2+\omega\}.
\end{aligned}
$$
For $m\ge1$, this gives $B\le K_1m^{1/6}$. Define $t_m=(k+\sigma^2h_m^2)/\omega$. The head drift and Lemma~\ref{lem:glue} give
$$
\CP(\pi)\le\frac{2\sigma^2}{\omega}
+(1+t_m)\CP(\pi_{H_B}).
$$
Consequently,
\begin{equation}
\label{eq:global_depth_full}
\begin{aligned}
\CP(\pi)\le{}&\sigma^2+\frac{2\sigma^2}{\omega}+\sigma^2t_m\\
&+2\sigma^4(1+t_m)e^{j e_m}\Delta_{m,L}\\
&+(1+t_m)e^{j e_m}
\left\{\frac{j}{um}+\sigma^2j e_m\right\}.
\end{aligned}
\end{equation}

\paragraph{Width threshold.}
Take $m\ge K_0\alpha^{9L}$, with $K_0\ge1$ chosen below. Then $L\le\log m/(9\log \alpha)$, so
$$
1+\sqrt{\log(mL)}\le K_1m^{1/6}.
$$
Equation~\eqref{eq:global_depth_delta} therefore gives
$$
\Delta_{m,L}\le K_1\alpha^{3L}m^{-1/3}
\le K_1K_0^{-1/3},
$$
where we used $\alpha^{4L}/m\le \alpha^{3L}m^{-1/3}$. Enlarge $K_0$ to ensure $\Delta_{m,L}\le(2\sigma^2)^{-1}$. Similarly,~\eqref{eq:global_depth_gradient} gives
$$
\max_\ell G_\ell^2/m
\le K_1\alpha^{2L}m^{-2/3}
\le K_1K_0^{-2/3}\alpha^{-4L}.
$$
A further increase of $K_0$ ensures $G_\ell^2/m\le4\zeta q_*$ for every hidden layer.

It remains to verify $p_G\ge1/2$. Set
$$
c_G=\frac{(\alpha-r_0-2\sigma)^2}{2\sigma^2}>0.
$$
For the chosen radius, $t=(B/\sigma-\sqrt k)^2/2\ge\log m+\omega$. Lemma~\ref{lem:mass} thus gives
$$
1-p_G\le e^{A_n}
\{e^{-m^{1/3}}+(L-2)e^{-c_Gm}\}.
$$
Since $L\le\log m/(9\log \alpha)$, we can enlarge $K_0$ once more to ensure $p_G\ge1/2$ and $j e_m\le1$, uniformly in $L$.

Finally, $t_m\le K_1m^{-1/3}$, while
$$
\frac{j}{um}+\sigma^2j e_m\le K_1m^{-1/3}.
$$
Substitution in~\eqref{eq:global_depth_full} proves
$$
\CP(\pi)\le\sigma^2+K \alpha^{3L}m^{-1/3}
\qquad(m\ge K_0\alpha^{9L}),
$$
with $K,K_0$ independent of $L,m$.
\end{proof}

At fixed depth, the upper PI bound has a matching Gaussian limit. A linear observable in one head column supplies the lower comparison, because its likelihood derivative is of order $m^{-1/2}$.

\begin{lemma}[Matching global gap test]
Under the same assumptions, at fixed depth,
\[
 \lambda_P(\pi)\le\sigma^{-2}+O(m^{-1/2}),\qquad
 \lambda_P(\pi)\longrightarrow\sigma^{-2}.
\]
\end{lemma}
\begin{proof}
For a unit vector $v\in\R^k$, let $x=v^Ta_j$, $x_0=v^Ta_{j,0}$, and $s^2=\Var_\pi(x)$. Integration by parts gives
\[
 1=s^2/\sigma^2+\operatorname{Cov}_\pi(x,\partial_x\V),\qquad
 |1-s^2/\sigma^2|\le h_ms.
\]
Gaussian domination and the bounded head derivative justify cutoff removal. Hence
\[
 s\ge\frac{\sqrt{\sigma^4h_m^2+4\sigma^2}-\sigma^2h_m}{2}
 =\sigma+O(m^{-1/2}).
\]
Testing PI with $x$, whose gradient has norm one, gives the upper gap bound. Combine with Theorem~\ref{thm:global}.
\end{proof}

\subsection{Supplementary conditional logarithmic Sobolev inequality}
This argument gives a width-uniform conditional LSI using the PI already established. The following calculation explains why the curvature-defect proof does not directly extend to entropy dissipation: differentiating the square root of the Fisher-information density produces an additional fourth-order term.

\begin{lemma}[Fourth-order term in entropy dissipation]\label{lem:entropy-fourth-order}
For a smooth positive density $r$, put $g=\sqrt r\,|\nabla\log r|$. Then, with regularization at zeros of $\nabla\log r$,
\[
 |\nabla g|^2\le2r\|\nabla^2\log r\|_{\HS}^2
 +\tfrac12r|\nabla\log r|^4.
\]
Even for a standard Gaussian reference, no constant $C$ bounds $\nu(r|\nabla\log r|^4)$ by $\nu(r\|\nabla^2\log r\|_{\HS}^2)+C\nu(r|\nabla\log r|^2)$ for all $r$.
\end{lemma}
\begin{proof}
Differentiate $g$ and apply $|u+v|^2\le2|u|^2+2|v|^2$, using $|\nabla|\nabla\log r||\le\|\nabla^2\log r\|_{\HS}$. For $\nu=N(0,I)$ and $r(x)=e^{t x_1-t^2/2}$, the Hessian term is zero, while the fourth-order and Fisher information terms are $t^4$ and $t^2$, respectively. Let $t\to\infty$.
\end{proof}

\begin{rem}[LSI argument]
Lemma~\ref{lem:entropy-fourth-order} concerns the curvature-defect argument for entropy dissipation on a fixed domain. This is separate from the difficulty of removing domain restrictions discussed in Remark~\ref{rem:pi_lsi_globalization}. The conditional LSI below uses exponential moments of the loss gradient to obtain a defective entropy bound, then combines it with the conditional PI.
\end{rem}

\begin{lemma}[Entropy transfer with random gradient]\label{lem:random-gradient}
Let $\CLS(\mu)\le\sigma^2$, let $V\ge0$, and write $d\nu=Z^{-1}e^{-V}d\mu$. If the displayed moments are finite, then
\[
 \Ent_\nu(g^2)\le8\sigma^2\nu(|\nabla g|^2)+b_V\nu(g^2),
\]
where
\[
 b_V=\frac12\left[\log\nu(e^{4\sigma^2|\nabla V|^2})
 +\log\nu(e^{4V})\right].
\]
\end{lemma}
\begin{proof}
The reference LSI and $\log Z\le0$ give
\[
 \Ent_\nu(g^2)\le4\sigma^2\nu(|\nabla g|^2)
 +\sigma^2\nu(g^2|\nabla V|^2)+\nu(g^2V).
\]
Entropy duality with parameters $4\sigma^2$ and $4$ absorbs one quarter of the entropy for each of the last two terms. Divide the remaining inequality by one half.
\end{proof}

The deterministic gradient bound in Lemma~\ref{lem:gradient} grows with the head cutoff. Retaining $\norm A_F^2/m$ inside the exponential moment avoids this growth, since its Gaussian exponential moments stay bounded as width increases. This improves the conditional LSI consequence of Theorem~\ref{thm:local} to a width-uniform bound.

\begin{theorem}[Width-uniform deep conditional LSI]\label{prop:deep-lsi}
For each fixed $L\ge2$ and fixed $\alpha$, assume $p_G\ge p_0>0$. On $G_{B,\alpha}$, retain the random head norm in the gradient estimate:
\[
 |\nabla\V|^2\le2\tau^2n^2
 \left(1+X^2\sum_{r=0}^{h-1}\alpha^{2r}\frac{\norm A_F^2}{m}\right).
\]
For $\nu=\pi(\cdot\mid G_{B,\alpha})$, set
\[
 b_L=\frac12\left[\log\nu(e^{4\sigma^2|\nabla\V|^2})
 +\log\nu(e^{4\V})\right].
\]
Then $b_L=O_L(1)$ for sufficiently large $m$. Whenever $\Delta_{m,L}<\sigma^{-2}$,
\[
 \boxed{\CLS(\nu)\le4\sigma^2+
 \frac{b_L+2}{2(\sigma^{-2}-\Delta_{m,L})}.}
\]
In particular, choose radii $B_m=O(\sqrt{\log m})$ such that $p_G\ge p_0$ for all sufficiently large $m$, and write $b_L(m)$ for the moment bound at width $m$. Choose $m_0$ so that $\Delta_{m,L}\le(2\sigma^2)^{-1}$ for $m\ge m_0$. Then
$$
\CLS(\nu)\le C_L
:=\sigma^2\left(6+\sup_{m\ge m_0}b_L(m)\right)<\infty.
$$
The constant $C_L$ may depend on the fixed data, depth, cutoff, loss scale, prior variance, activation and prior-mean bounds, and $p_0$, but not on $m$.
\end{theorem}
\begin{proof}
In the block-gradient proof, use $\norm A_{\op}\le\norm A_F$ instead of $\norm A_{\op}\le b\sqrt m$. The Gaussian quadratic moment is explicit:
\[
\begin{aligned}
 \gamma(e^{s\norm A_F^2/m})
 ={}&(1-2\sigma^2s/m)^{-km/2}\\
 &\times\exp\left(\frac{s\norm{A_0}_F^2/m}{1-2\sigma^2s/m}\right).
\end{aligned}
\]
for $m>2\sigma^2s$. Its logarithm is bounded for each fixed $s$, since $\norm{A_0}_F^2\le k a_0^2$. Domination by $e^{A_n}\gamma/p_G$ gives the first moment defining $b_L$; Lemma~\ref{lem:mass} gives the second. Apply Lemma~\ref{lem:random-gradient} to the truncated Gaussian reference, and use Rothaus' inequality with Theorem~\ref{thm:local}.
\end{proof}

\subsection{Linear and two-layer conditional inequalities}\label{sec:linear-two-local}
For $L=1$, interpret the predictor as affine in the parameters, with no hidden layers. Convex cross-entropy gives $\nabla^2U\succeq\sigma^{-2}I$ globally. Consequently $\CP,\CLS\le\sigma^2$ globally and on every convex conditioning domain.

For two layers, the neuronwise structure also gives a direct pointwise curvature bound on $H_B$. This supplies both inequalities on the cylinder without using the random curvature envelope.

\begin{lemma}[Two-layer conditional curvature]\label{prop:two-local}
For $L=2$, let $c_1=\norm{\phi'}_\infty\le1$. On the centered cylinder $H_B$, put $b=B+b_0$ and
\[
\begin{aligned}
 u&=\sqrt2\tau n c_1X,\qquad v_b=\sqrt2\tau n c_\phi bX^2,\\
 \varepsilon_b&=\frac{v_b+\sqrt{v_b^2+4u^2}}{2\sqrt m}.
\end{aligned}
\]
Then $\nabla^2U\succeq(\sigma^{-2}-\varepsilon_b)I$. If $\varepsilon_b<\sigma^{-2}$,
\[
 \CP(\pi_{H_B}),\CLS(\pi_{H_B})
 \le(\sigma^{-2}-\varepsilon_b)^{-1}.
\]
On the uncentered cylinder $\{\max_j\norm{a_j}\le B\}$, replace $b$ by $B$.
\end{lemma}
\begin{proof}
For a parameter direction $(\dot A,\dot W)$, the potentially negative loss-Hessian contribution is bounded in absolute value by
\[
 m^{-1/2}\left[2u\sum_j\norm{\dot a_j}\norm{\dot w_j}
 +v_b\sum_j\norm{\dot w_j}^2\right].
\]
The largest eigenvalue of $\left(\begin{smallmatrix}0&u\\u&v_b\end{smallmatrix}\right)$ is $(v_b+\sqrt{v_b^2+4u^2})/2$. Sum over neurons, discard the positive prediction-space loss Hessian, and apply convex-domain Bakry--\'Emery.
\end{proof}
For fixed data and a high-mass cutoff $B=O(\sqrt{\log(m/\delta)})$, the correction is $O(\sqrt{\log(m/\delta)/m})$, provided this tends to zero. At fixed $B$, it is $O(m^{-1/2})$.

\subsection{Alternative global two-layer LSI via clipping}
We give an alternative global LSI by clipping the head in the predictor, transferring the resulting LSI back to the original posterior with a small entropy defect, and removing that defect using the global PI. This route yields a bound approaching $2\sigma^2$. Appendix~\ref{app:lsi} proves the sharper bound approaching $\sigma^2$ by exploiting conditional head convexity.

We compare $\pi$ with a posterior whose predictor has smoothly truncated head columns. Choose a fixed smooth map $q:\R^k\to\R^k$ that equals the identity on the unit ball and vanishes outside the ball of radius two; for example $q(a)=\chi(\norm a^2)a$ with a smooth cutoff $\chi$. Define $q_B(a)=Bq(a/B)$ for $B\ge1$. Then
\[
\begin{aligned}
 q_B(a)&=a\quad(\norm a\le B),\\
 \norm{q_B(a)}&\le C\min\{\norm a,B\},\\
 \norm{Dq_B}&\le C,\qquad \norm{D^2q_B}\le C/B.
\end{aligned}
\]
Set
\[
\begin{aligned}
 f_{B,i}&=m^{-1/2}\sum_jq_B(a_j)\phi(w_j^Tx_i),\\
 \V_B&=\tau\sum_i\ell_i(f_{B,i}),\\
 d\mu_B&=Z_B^{-1}e^{-\V_B}d\gamma.
\end{aligned}
\]

\begin{lemma}[Uniform curvature after clipping]\label{lem:clip-curvature}
For $L=2$,
\[
 \norm{D^2f_{B,i}}_{\rm bil}\le\frac{C(B+1+B^{-1})}{\sqrt m}.
\]
Consequently, for $B\ge1$ and $C(B+1)/\sqrt m<\sigma^{-2}$,
\[
 \CLS(\mu_B)\le C_0:=
 [\sigma^{-2}-C(B+1)/\sqrt m]^{-1}.
\]
\end{lemma}
\begin{proof}
The predictor is a sum of neuronwise maps. Within each neuron, its second derivative has head--head bound $C/B$, head--hidden bound $Cc_1X$, and hidden--hidden bound $CBc_\phi X^2$. Cauchy--Schwarz over the neuron blocks bounds the bilinear norm by their maximum block bound, times $m^{-1/2}$, with no factor $m$. The loss Hessian consists of a positive semidefinite term and contraction against $\tau\nabla\ell_i$. Global Bakry--\'Emery gives the result.
\end{proof}

Clipping changes the loss only when a head column leaves the cutoff ball. The next lemma quantifies both the density perturbation and its gradient, since both appear when transferring the LSI from $\mu_B$ to $\pi$.

\begin{lemma}[Clipping defects and exponentially small moments]\label{lem:clip-moments}
Let $R=\V-\V_B$ and $E_B=\{\max_j\norm{a_j}>B\}$. Both $R$ and $\nabla R$ vanish on $E_B^c$, and
\begin{align}
 |R|&\le\frac C{\sqrt m}\sum_j\norm{a_j}\mathbf1_{\{\norm{a_j}>B\}},\label{eq:R-envelope}\\
 |\nabla R|^2&\le C\left(1+m^{-1}\sum_j\norm{a_j}^2\right)\mathbf1_{E_B}.\label{eq:gradR-envelope}
\end{align}
There are fixed $c_0,c,C,K>0$ such that, if
\[
 K\log m\le B^2\le\sqrt m,\qquad t=c_0B^2,
\]
then, for sufficiently large $m$,
\[
 \delta_B:=\log\pi(e^{t|R|})+\log\pi(e^{t|\nabla R|^2})
 \le Cm e^{-cB^2}.
\]
All constants allow the uniformly bounded nonzero prior centers assumed in this appendix.
\end{lemma}
\begin{proof}
The loss is Lipschitz and $q_B(a)=a$ on the ball, giving \eqref{eq:R-envelope}. The global two-layer gradient bounds for $\V$ and $\V_B$ are each at most $C(1+\norm A_F^2/m)$ after squaring. Their difference is zero outside $E_B$, proving \eqref{eq:gradR-envelope}. Write $Y_j=\norm{a_j}$ under the independent prior heads. Uniformly bounded centers imply $\gamma(Y_j>B)\le Ce^{-cB^2}$. For $\lambda=Ct/\sqrt m$, the imposed range makes $\lambda$ bounded. Gaussian integration gives
\[
 \gamma(e^{\lambda Y_j\mathbf1_{Y_j>B}})-1
 =\gamma[(e^{\lambda Y_j}-1)\mathbf1_{Y_j>B}]
 \le Ce^{-cB^2}.
\]
Independence and \eqref{eq:R-envelope} imply $\gamma(e^{t|R|})-1\le\exp(Cm e^{-cB^2})-1\le Cm e^{-cB^2}$ after increasing $K$. For $J=C(1+\norm A_F^2/m)$, Cauchy--Schwarz gives
\[
\begin{aligned}
 \gamma(e^{t|\nabla R|^2}-1)
 &\le\gamma(e^{2tJ})^{1/2}\gamma(E_B)^{1/2}\\
 &\le C\sqrt m\,e^{-cB^2+C't}.
\end{aligned}
\]
Here the Gaussian quadratic formula gives $\log\gamma(e^{2tJ})\le C't$ because $t=O(\sqrt m)$ and $\norm{A_0}_F^2=O(1)$. Choose $c_0$ sufficiently small to retain a negative multiple of $B^2$. Finally apply domination to the nonnegative quantities $e^{t|R|}-1$ and $e^{t|\nabla R|^2}-1$, rather than to the full exponentials. This avoids a fixed additive $A_n$ in their logarithms and proves the claim.
\end{proof}

\begin{theorem}[Near-Gaussian defective global LSI for two layers]\label{thm:sharp-defective}
For $L=2$, with the preceding range of $B$,
\[
\begin{aligned}
 \Ent_\pi(f^2)\le{}&
 2\left[\sigma^2+O\left(\frac B{\sqrt m}+\frac1B\right)\right]
 \pi(|\nabla f|^2)\\
 &+Cm e^{-cB^2}\pi(f^2).
\end{aligned}
\]
In particular, $B=m^{1/4}$ gives
\[
\begin{aligned}
 \Ent_\pi(f^2)\le{}&2[\sigma^2+O(m^{-1/4})]\pi(|\nabla f|^2)\\
 &+Ce^{-c\sqrt m}\pi(f^2).
\end{aligned}
\]
\end{theorem}
\begin{proof}
Write $d\pi=Z_R^{-1}e^{-R}d\mu_B$. Applying the reference LSI to $f e^{-R/2}/\sqrt{Z_R}$ and using Young's inequality gives, for $\epsilon>0$,
\begin{align*}
 \Ent_\pi(f^2)\le{}&2C_0(1+\epsilon)\pi(|\nabla f|^2)\\
 &+c_\epsilon\pi(f^2|\nabla R|^2)\\
 &+\pi(f^2|R|)+|\log Z_R|\pi(f^2),
\end{align*}
Here $c_\epsilon=\tfrac12C_0(1+\epsilon^{-1})$. Since $\pi(e^R)=Z_R^{-1}$, Jensen and H\"older, for $t\ge1$, imply
\[
 |\log Z_R|\le t^{-1}\log\pi(e^{t|R|}).
\]
Entropy duality applied to both random defects with parameter $t$ therefore gives
\[
\begin{aligned}
 (1-\theta_B)\Ent_\pi(f^2)
 \le{}&2C_0(1+\epsilon)\pi(|\nabla f|^2)\\
 &+\frac{c_\epsilon+2}{t}\delta_B\pi(f^2),\\
 \theta_B={}&\frac{c_\epsilon+1}{t}.
\end{aligned}
\]
Take $\epsilon=B^{-1}$ and $t=c_0B^2$. Then $\theta_B=O(B^{-1})<1$, $C_0=\sigma^2+O(B/\sqrt m)$, and the defective term remains $Cm e^{-cB^2}$. Finally absorb the polynomial $m$ into the exponential when $B^2=\sqrt m$.
\end{proof}

The preceding defective inequality has a gradient coefficient approaching the prior scale. Removing its additive defect with Rothaus' inequality also introduces the global Poincar\'e constant, producing the factor $2$ in the following bound.

\begin{theorem}[Global two-layer LSI]\label{cor:two-global-lsi}
Under the assumptions of Theorem~\ref{thm:global}, at $L=2$,
\[
 \boxed{\CLS(\pi)\le2\sigma^2+O(m^{-1/4}).}
\]
\end{theorem}
\begin{proof}
If $\Ent_\pi(f^2)\le2C\pi(|\nabla f|^2)+D\pi(f^2)$, Rothaus' inequality gives
\[
 \CLS(\pi)\le C+(1+D/2)\CP(\pi).
\]
Apply Theorem~\ref{thm:sharp-defective} and $\CP(\pi)\le\sigma^2+O(m^{-1/3})$.
\end{proof}

\begin{table*}[t]
\centering\small
\setlength{\tabcolsep}{3pt}
\caption{Bounds obtained by the curvature-defect, Gaussian-drift, and clipping arguments for fixed data, depth, loss scale, and prior variance, with $\epsilon_m=\sqrt{\log m/m}$ and width-independent bounds on $a_0,r_0$. Conditional bounds use $B\asymp\sqrt{\log m}$ and, for $L>2$, fixed $\alpha>r_0+2\sigma$. The global PI proof uses $\alpha>r_0+3.853\sigma$. A dash indicates that this argument gives no bound. The sharper LSI bounds in Section~\ref{sec:lsi} are proved separately in Appendix~\ref{app:lsi}.}
\label{tab:cylinder-bounds}
\begin{tabular}{@{}p{.08\textwidth}p{.14\textwidth}p{.18\textwidth}p{.17\textwidth}p{.18\textwidth}p{.17\textwidth}@{}}
\toprule
Depth & Conditional\newline domain & Conditional $\CP$ & Conditional $\CLS$ & Global $\CP$ & Global $\CLS$\\
\midrule
$L=1$ & Convex & $\le\sigma^2$ & $\le\sigma^2$ & $\le\sigma^2$ & $\le\sigma^2$\\[4pt]
$L=2$ & $H_B$ & $\le\sigma^2+O(\epsilon_m)$ & $\le\sigma^2+O(\epsilon_m)$ & $\le\sigma^2+O(m^{-1/3})$ & $\le2\sigma^2+O(m^{-1/4})$\\[4pt]
$L>2$ & $G_{B,\alpha}$ & $\le\sigma^2+O_L(\epsilon_m)$ & $\le C_L$ & $\le\sigma^2+O_L(m^{-1/3})$ & ---\\
\bottomrule
\end{tabular}
\par\smallskip
The linear case is given in Section~\ref{sec:linear-two-local}; the two-layer conditional bounds in Lemma~\ref{prop:two-local}; the deep conditional bounds in Theorem~\ref{thm:local} and Theorem~\ref{prop:deep-lsi}; and the global bounds in Theorem~\ref{thm:global} and Theorem~\ref{cor:two-global-lsi}.
\end{table*}

\section{Appendix: Logarithmic Sobolev bounds}
\label{app:lsi}

We give finite-width versions of the shallow global and deep local LSI bounds. The proof has three parts: estimates for the conditional law of the head $A$, curvature of the hidden-weight marginal, and control of the conditional score needed to combine the two inequalities. The posterior-mass estimate is proved separately in Section~\ref{app:spectral_mass}.

\subsection{Finite-width statements}
\label{app:lsi_statements}

We use the model in Section~\ref{sec:model}, with integers $n,m,d\ge1$, $C\ge2$, and $L\ge2$. Put $h=L-1$, $W=(W_1,\ldots,W_h)$, and $\theta=(A,W)$. The features and outputs in~\eqref{eq:network} are
\begin{equation}
\label{eq:app_lsi_network}
\begin{aligned}
H_1(x)&=\phi(W_1x),\\
H_\ell(x)&=\phi\!\left(W_\ell H_{\ell-1}(x)/\sqrt m\right),\quad 2\le\ell\le h,\\
f_\theta(x)&=AH_h(x)/\sqrt m.
\end{aligned}
\end{equation}
Using the same centered-logit basis $Q$, write the individual and summed losses as
\begin{equation}
\label{eq:app_lsi_loss}
\begin{aligned}
\ell_i(z)&=\log\!\left(\sum_{c=1}^C e^{(Qz)_c}\right)-(Qz)_{y_i},\\
V(\theta)&=\sum_{i=1}^n\ell_i(f_\theta(x_i)).
\end{aligned}
\end{equation}
For $\sigma>0$ and a deterministic center $\theta_0=(A_0,W_{1,0},\ldots,W_{h,0})$, the prior and posterior are
\begin{equation}
\label{eq:app_lsi_posterior}
\begin{aligned}
\gamma&=N(\theta_0,\sigma^2I),\\
\pi(d\theta)&=Z^{-1}e^{-V(\theta)}\gamma(d\theta),\qquad Z=\E_\gamma e^{-V}.
\end{aligned}
\end{equation}

Throughout this section, the hypotheses of Assumptions~\ref{ass:activation}, \ref{ass:prior_means}, and~\ref{ass:inputs} are the following explicit bounds:
\begin{equation}
\label{eq:app_lsi_assumptions}
\begin{gathered}
\phi\in C^2(\R),\qquad |\phi|,|\phi'|\le1,\qquad |\phi''|\le c_\phi,\\
\max_{1\le i\le n}\norm{x_i}\le \bar{x},\qquad \norm{A_0}_{\op}\le a_0,\\
\max_{2\le\ell\le h}\norm{W_{\ell,0}/\sqrt m}_{\op}\le r_0\qquad (L>2).
\end{gathered}
\end{equation}
Here $\bar{x},c_\phi,a_0,r_0$ are nonnegative finite constants, uniform in $n,m$ when these vary. The first-layer center $W_{1,0}$ is arbitrary.

For $L>2$, fix $a>r_0+2\sigma$ and recall the domain and conditioned posterior
\begin{equation}
\label{eq:app_lsi_domain}
\begin{aligned}
\mathcal G_a&=\bigcap_{\ell=2}^{h}\{\theta:\norm{W_\ell/\sqrt m}_{\op}\le a\},\\
\nu_a&=\pi(\cdot\mid\mathcal G_a).
\end{aligned}
\end{equation}
The head $A$ and the first layer remain unrestricted. For statements applying to both depths, write $\nu=\pi$ when $L=2$ and $\nu=\nu_a$ when $L>2$. All gradients are with respect to the parameter entries; on $\mathcal G_a$ we use the ambient gradient restricted to this domain. With the conventions of Definitions~\ref{def:pi} and~\ref{def:lsi}, our constants satisfy
\begin{equation}
\label{eq:app_lsi_convention}
\begin{aligned}
\Var_\nu(g)&\le\CP(\nu)\E_\nu\norm{\nabla g}^2,\\
\Ent_\nu(g^2)&\le2\CLS(\nu)\E_\nu\norm{\nabla g}^2.
\end{aligned}
\end{equation}

\paragraph{Coefficients in the finite-width bound.}
The constant $c_0$ controls the conditional evidence, while $K$ bounds second moments of fixed projections of the head, as shown in Lemma~\ref{lem:app_lsi_head}. Set
\begin{equation}
\label{eq:app_lsi_head_constants}
\begin{aligned}
c_0&=\log C+\sqrt2\,a_0+\frac{\sigma^2(C-1)}{2C},\\
K&=\sigma^2(C-1)+\bigl(a_0+\sigma\sqrt{2nc_0}\bigr)^2.
\end{aligned}
\end{equation}
For two layers, the data enter through
\begin{equation}
\label{eq:app_lsi_chi}
\chi=\lambda_{\max}\!\left(\sum_{i=1}^n x_ix_i^\top\right)\le n\bar{x}^2.
\end{equation}
For greater depth, three coefficients record propagation through the hidden layers. Set $w_1=\bar{x}$ and $w_j=1$ for $2\le j\le h$, and define $p_\ell\in\R^h$ by
\begin{equation}
\label{eq:app_lsi_p_vectors}
(p_\ell)_j=\begin{cases}w_j a^{\ell-j},&j\le\ell,\\0,&j>\ell.\end{cases}
\end{equation}
Let $R\in\R^{h\times h}$ be symmetric, with zero diagonal and entries $R_{j\ell}=R_{\ell j}=w_j a^{h-1-j}$ for $j<\ell$. Put
\begin{equation}
\label{eq:app_lsi_derivative_constants}
\begin{aligned}
S&=\norm{p_h},\\
T&=c_\phi\lambda_{\max}\!\left(\sum_{\ell=1}^h a^{h-\ell}p_\ell p_\ell^\top\right),\\
T_{\mathrm{mix}}&=\lambda_{\max}(R).
\end{aligned}
\end{equation}
The coefficient $S$ bounds first feature variations. The coefficients $T$ and $T_{\mathrm{mix}}$ bound second variations arising from activation curvature and mixed weight--feature derivatives, respectively. They depend only on $L,a,\bar{x},c_\phi$. Section~\ref{app:feature_derivatives} proves the derivative estimates they encode.

The coefficient $G$ collects squared first-derivative bounds over the observations, while $\kappa_1$ and $\kappa_2$ collect marginal-curvature losses at orders $m^{-1/2}$ and $m^{-1}$. For $L=2$, set
\begin{equation}
\label{eq:app_lsi_shallow_coefficients}
\begin{aligned}
G&=\chi,\\
\kappa_1&=2c_\phi a_0\chi,\\
\kappa_2&=\sigma^2\chi\bigl[c_\phi(1+2n)+2n\bigr].
\end{aligned}
\end{equation}
For $L>2$, set
\begin{equation}
\label{eq:app_lsi_deep_coefficients}
\begin{aligned}
G&=nS^2,\\
\kappa_1&=n\sqrt{2K}\,T,\\
\kappa_2&=n\sqrt{2K}\,T_{\mathrm{mix}}+2\sigma^2nG.
\end{aligned}
\end{equation}

\begin{theorem}[Finite-width shallow global and deep local LSI]
\label{thm:app_lsi_finite}
Consider the network and posterior in~\eqref{eq:app_lsi_network}--\eqref{eq:app_lsi_posterior}, under the bounds~\eqref{eq:app_lsi_assumptions}. For $L=2$, take $\nu=\pi$ and the coefficients in~\eqref{eq:app_lsi_shallow_coefficients}. For $L>2$, fix $a>r_0+2\sigma$, take $\nu=\nu_a$ from~\eqref{eq:app_lsi_domain}, and use~\eqref{eq:app_lsi_derivative_constants} and~\eqref{eq:app_lsi_deep_coefficients}. Define
\begin{equation}
\label{eq:app_lsi_rho_q}
\begin{aligned}
\delta_m&=\frac{\kappa_1}{\sqrt m}+\frac{\kappa_2}{m},\qquad \rho=\sigma^{-2}-\delta_m,\\
q&=\frac{G}{m}\left(2\sigma^2n+\frac K2\right).
\end{aligned}
\end{equation}

Here $\delta_m$ bounds the loss of hidden-marginal curvature, $\rho$ is the remaining curvature, and $q$ controls how the conditional head distribution varies with the hidden weights.

If $\rho>0$, then
\begin{equation}
\label{eq:app_lsi_exact_bound}
\frac{1}{\sigma^{-2}+n/(2m)}\le\CP(\nu)\le\CLS(\nu)\le\lambda_{\max}(M),
\end{equation}
where
\begin{equation}
\label{eq:app_lsi_matrix}
M=\begin{pmatrix}
\sigma^2(1+q/\rho)&\sigma\sqrt q/\rho\\
\sigma\sqrt q/\rho&1/\rho
\end{pmatrix}.
\end{equation}
The condition $\rho>0$ is exactly
\begin{equation}
\label{eq:app_lsi_positive_width}
m>\frac14\left(\sigma^2\kappa_1+\sqrt{\sigma^4\kappa_1^2+4\sigma^2\kappa_2}\right)^2.
\end{equation}
In particular, the explicit sufficient condition
\begin{equation}
\label{eq:app_lsi_half_width}
m\ge\max\{1,16\sigma^4\kappa_1^2,4\sigma^2\kappa_2\}
\end{equation}
ensures $\rho\ge(2\sigma^2)^{-1}$ and gives
\begin{equation}
\label{eq:app_lsi_additive_bound}
\CLS(\nu)\le\sigma^2+2\sigma^4(\delta_m+q)+2\sigma^3\sqrt q.
\end{equation}
The first-layer center is arbitrary, and no bound in the statement depends on the ambient parameter dimension.
\end{theorem}

For fixed $C,\sigma,\bar{x},c_\phi,a_0$, the shallow coefficients satisfy $K=O(n)$, $G=O(n)$, $\kappa_1=O(n)$, and $\kappa_2=O(n^2)$. At fixed $L,a$, the deep coefficients satisfy $\kappa_1=O_{L,a}(n^{3/2})$ and $\kappa_2=O_{L,a}(n^2)$. Thus~\eqref{eq:app_lsi_half_width} makes the $n^2$ and $n^3$ sufficient-width scales explicit, and~\eqref{eq:app_lsi_additive_bound} yields the respective errors in Theorems~\ref{thm:shallow_lsi} and~\ref{thm:deep_lsi}.

\begin{corollary}[Width for a prescribed LSI accuracy]
\label{cor:app_lsi_accuracy}
In the setting of Theorem~\ref{thm:app_lsi_finite}, fix $0<\varepsilon\le1$. The bound $\CLS(\nu)\le(1+\varepsilon)\sigma^2$ holds whenever
\begin{equation}
\label{eq:app_lsi_accuracy_width}
\begin{split}
m\ge\max\Biggl\{&1,\frac{256\sigma^4\kappa_1^2}{\varepsilon^2},\frac{16\sigma^2\kappa_2}{\varepsilon},\\
&\frac{16\sigma^2G}{\varepsilon^2}\left(2\sigma^2n+\frac K2\right)\Biggr\}.
\end{split}
\end{equation}
For $L>2$ and any $0<\eta<1$, imposing also~\eqref{eq:app_lsi_mass_width} guarantees posterior mass at least $1-\eta$.
\end{corollary}

The proofs of the theorem and corollary are completed in Section~\ref{app:lsi_assembly}, after establishing the conditional, marginal, and score estimates.

\subsection{Conditional-head bounds}
\label{app:lsi_head}

Vectorize the head as $\alpha=\operatorname{vec}(A)$ and its prior mean as $\alpha_0=\operatorname{vec}(A_0)$. For fixed $W$, define the linear map $\Phi(W):\R^{(C-1)m}\to\R^{n(C-1)}$ by
\begin{equation}
\label{eq:app_lsi_feature_map}
\Phi(W)\alpha=\bigl(AH_h(x_i)/\sqrt m\bigr)_{i=1}^n.
\end{equation}
For a stacked prediction vector $z=(z_1,\ldots,z_n)$, put $\ell(z)=\sum_i\ell_i(z_i)$. Thus $V(A,W)=\ell(\Phi(W)\alpha)$. Let $\gamma_A=N(\alpha_0,\sigma^2I)$ and define
\begin{equation}
\label{eq:app_lsi_head_conditional}
\begin{aligned}
Z_W&=\E_{\gamma_A}e^{-V(A,W)},\\
\mu_W(d\alpha)&=Z_W^{-1}e^{-V(A,W)}\gamma_A(d\alpha).
\end{aligned}
\end{equation}
Because the domain restricts only $W$, this is the conditional law of the head under both $\pi$ and $\nu_a$. Below, $r_i=\nabla\ell_i(f_\theta(x_i))$ and $\Lambda_i=\nabla^2\ell_i(f_\theta(x_i))$. Their stacked versions are $r=\nabla\ell(\Phi\alpha)$ and $\Lambda=\nabla^2\ell(\Phi\alpha)$, with $\Lambda$ block diagonal.

The next lemma supplies the conditional LSI used in entropy assembly and the head moments needed for both marginal curvature and score control. The projected bound for $Az$ estimates the head only along the feature-derivative directions that enter those arguments.

\begin{lemma}[Loss derivatives and head moments]
\label{lem:app_lsi_head}
For every hidden-weight configuration $W$,
\begin{equation}
\label{eq:app_lsi_loss_bounds}
\begin{gathered}
\norm{r_i}\le\sqrt2,\qquad \norm r^2\le2n,\\
0\preceq\Lambda\preceq\tfrac12 I,\qquad \operatorname{tr}\Lambda_i\le1.
\end{gathered}
\end{equation}
With $c_0,K$ from~\eqref{eq:app_lsi_head_constants},
\begin{equation}
\label{eq:app_lsi_head_bounds}
\begin{gathered}
Z_W\ge e^{-nc_0},\qquad \CLS(\mu_W)\le\sigma^2,\\
\Cov_{\mu_W}(\alpha)\preceq\sigma^2I,\\
\E_{\mu_W}\norm{Az}^2\le K\norm z^2\qquad(z\in\R^m).
\end{gathered}
\end{equation}
\end{lemma}

\begin{proof}
Let $p_i$ be the vector of softmax probabilities at observation $i$, and let $e_{y_i}$ be the corresponding coordinate vector in $\R^C$. Then
\[
r_i=Q^\top(p_i-e_{y_i}),\qquad
\Lambda_i=Q^\top\bigl(\operatorname{diag}(p_i)-p_ip_i^\top\bigr)Q.
\]
The gradient bound follows from $\norm{p_i-e_{y_i}}\le\sqrt2$. The uncentered Hessian is positive semidefinite; its absolute row sums are $2p_{i,c}(1-p_{i,c})\le1/2$, and its trace is $1-\norm{p_i}^2\le1$. Restriction by the isometry $Q$ gives~\eqref{eq:app_lsi_loss_bounds}.

Write $u_i=H_h(x_i)/\sqrt m$, so $\norm{u_i}\le1$. Under $\gamma_A$, the vector $QAu_i$ has mean $QA_0u_i$ and every coordinate has variance $\sigma^2\norm{u_i}^2(C-1)/C$. Jensen's inequality for the logarithm and the Gaussian exponential-moment formula give
\[
\E_{\gamma_A}\ell_i(Au_i)\le\ell_i(A_0u_i)+\frac{\sigma^2(C-1)}{2C}\le c_0.
\]
The last inequality uses the $\sqrt2$-Lipschitz property of $\ell_i$ and $\ell_i(0)=\log C$. Summing and applying Jensen again yields $Z_W\ge\exp(-\E_{\gamma_A}V)\ge e^{-nc_0}$.

The conditional negative log-density has Hessian
\begin{equation}
\label{eq:app_lsi_head_hessian}
H_A=\sigma^{-2}I+\Phi^\top\Lambda\Phi\succeq\sigma^{-2}I.
\end{equation}
The Bakry--\'Emery criterion gives the conditional LSI \citep[Section~5.7]{bakry2014analysis}. Applying the Brascamp--Lieb variance inequality to linear functions gives the covariance bound \citep[Eq.~(1.7)]{corderoerausquin2017transport}.

The covariance bound controls fluctuations around the conditional mean. To obtain the projected second-moment bound, we also need to control the displacement of this mean from its prior value. We use $\operatorname{KL}(\xi\|\mu)=\int\log(d\xi/d\mu)\,d\xi$ for relative entropy when $\xi\ll\mu$. Since $V\ge0$,
\[
\operatorname{KL}(\mu_W\|\gamma_A)=-\E_{\mu_W}V-\log Z_W\le nc_0.
\]
The entropy variational inequality and the Gaussian moment-generating function give, for every vector $b$,
\[
b^\top(\E_{\mu_W}\alpha-\alpha_0)\le nc_0+\frac{\sigma^2}{2}\norm b^2.
\]
Optimizing over $b$ shows that $\norm{\E_{\mu_W}A-A_0}_{\HS}^2\le2\sigma^2nc_0$, and hence $\norm{\E_{\mu_W}A}_{\op}\le a_0+\sigma\sqrt{2nc_0}$. For each fixed $z$, the covariance bound controls the variance of each of the $C-1$ coordinates of $Az$. Adding these variances to the squared mean proves the last bound in~\eqref{eq:app_lsi_head_bounds}.
\end{proof}

\subsection{Feature derivatives}
\label{app:feature_derivatives}

We next quantify how a perturbation of the hidden weights changes the features. First derivatives control both the interaction cost in the marginal-curvature estimate and the conditional score. Second derivatives enter the nonlinear-feature remainder. A hidden direction is a tuple $v=(v_1,\ldots,v_h)$, with $\norm v^2=\sum_\ell\norm{v_\ell}_{\HS}^2$. For a differentiable map, $D_WH[v]$ denotes the derivative of $H$ in direction $v$.

\begin{lemma}[First and second feature derivatives]
\label{lem:app_lsi_derivatives}
For either depth regime, throughout the relevant hidden domain,
\begin{equation}
\label{eq:app_lsi_first_derivative}
\sum_{i=1}^n\norm{D_WH_h(x_i)[v]}^2\le G\norm v^2.
\end{equation}
For $L>2$, the coefficients in~\eqref{eq:app_lsi_derivative_constants} also give, for every observation,
\begin{equation}
\label{eq:app_lsi_second_derivative}
\begin{aligned}
\norm{D_WH_h(x_i)[v]}&\le S\norm v,\\
\norm{D_W^2H_h(x_i)[v,v]}&\le\left(T+\frac{T_{\mathrm{mix}}}{\sqrt m}\right)\norm v^2.
\end{aligned}
\end{equation}
In particular, for $E_v=D_W\Phi(W)[v]$,
\begin{equation}
\label{eq:app_lsi_map_derivative}
\begin{aligned}
\norm{E_v}_{\op}^2&\le\frac Gm\norm v^2,\\
\E_{\mu_W}\norm{E_v\alpha}^2&\le\frac{GK}{m}\norm v^2.
\end{aligned}
\end{equation}
\end{lemma}

\begin{proof}
Fix an observation and put $t_\ell=\norm{v_\ell}_{\HS}$, $t=(t_1,\ldots,t_h)^\top$, and $u_\ell=D_WH_\ell(x_i)[v]$. Differentiating the network, using $|\phi'|\le1$ and $\norm{H_\ell(x_i)}\le\sqrt m$, gives
\[
\norm{u_1}\le \bar{x}t_1,\qquad \norm{u_\ell}\le t_\ell+a\norm{u_{\ell-1}}\le p_\ell^\top t.
\]
This proves the first bound in~\eqref{eq:app_lsi_second_derivative}. For $\eta_\ell=D_W^2H_\ell(x_i)[v,v]$, the chain rule gives
\[
\norm{\eta_1}\le c_\phi \bar{x}^2t_1^2,
\]
and, for $\ell\ge2$,
\[
\norm{\eta_\ell}\le c_\phi(p_\ell^\top t)^2+a\norm{\eta_{\ell-1}}+\frac{2t_\ell}{\sqrt m}\norm{u_{\ell-1}}.
\]
Iterating the recurrence yields
\[
\norm{\eta_h}\le c_\phi\sum_{\ell=1}^h a^{h-\ell}(p_\ell^\top t)^2+\frac1{\sqrt m}t^\top Rt.
\]
The definitions of $T,T_{\mathrm{mix}}$ prove the second bound. Summing the first bound over observations gives~\eqref{eq:app_lsi_first_derivative} with $G=nS^2$.

For $L=2$, let $v_j$ denote row $j$ of $v_1$. The magnitude of coordinate $j$ of $D_WH_1(x_i)[v]$ is at most $|v_jx_i|$. Therefore
\[
\sum_{i,j}|v_jx_i|^2\le\chi\norm v^2,
\]
which proves~\eqref{eq:app_lsi_first_derivative} with $G=\chi$. Finally, block $i$ of $E_v\alpha$ is $A D_WH_h(x_i)[v]/\sqrt m$. The operator bound follows by summing the squared feature norms, and the moment bound follows by applying Lemma~\ref{lem:app_lsi_head} to each feature-derivative vector.
\end{proof}

\subsection{Hidden-marginal curvature}
\label{app:lsi_curvature}

Let $\bar\nu$ be the distribution of $W$ under $\nu$, obtained by integrating out $A$. Thus
\begin{equation}
\label{eq:app_lsi_disintegration}
\nu(dA,dW)=\mu_W(dA)\,\bar\nu(dW).
\end{equation}
For $L=2$, $\bar\nu=\bar\pi$ is the unrestricted hidden marginal from the main text. For $L>2$, it is the hidden marginal of $\nu_a$, supported on the hidden part of $\mathcal G_a$. In the interior of its support, its negative log-density is, up to an additive constant,
\begin{equation}
\label{eq:app_lsi_marginal_potential}
\bar U(W)=\frac{\norm{W-W_0}^2}{2\sigma^2}-\log Z_W,
\end{equation}
where $W_0=(W_{1,0},\ldots,W_{h,0})$.

The proof first combines the positive loss curvature with the variance created by integrating out the head. It then bounds the remaining feature term using neuronwise integration by parts for two layers and projected head moments at greater depth.

\begin{lemma}[Retained marginal curvature]
\label{lem:app_lsi_curvature}
With $\rho$ from~\eqref{eq:app_lsi_rho_q}, the hidden marginal satisfies $\nabla_W^2\bar U\succeq\rho I$ throughout the interior of its support.
\end{lemma}

\begin{proof}
Fix a unit hidden direction $v$ and abbreviate $E=D_W\Phi[v]$, $F=D_W^2\Phi[v,v]$, and $z=E\alpha$. Differentiating under the head integral gives
\begin{equation}
\label{eq:app_lsi_curvature_identity}
\begin{aligned}
\partial_v^2\bar U
={}&\sigma^{-2}+\E_{\mu_W}\bigl[z^\top\Lambda z+r^\top F\alpha\bigr]\\
&-\Var_{\mu_W}(r^\top E\alpha).
\end{aligned}
\end{equation}
The derivatives are justified locally in $W$ by domination with Gaussian-integrable polynomials in $\alpha$: the loss gradient and Hessian are bounded, while $V$ and its first two hidden derivatives have at most quadratic growth in $\alpha$.

The head gradient of $r^\top E\alpha$ is $b=E^\top r+\Phi^\top\Lambda z$. Brascamp--Lieb with the full Hessian $H_A$ from~\eqref{eq:app_lsi_head_hessian} gives
\[
\Var_{\mu_W}(r^\top E\alpha)\le\E_{\mu_W}[b^\top H_A^{-1}b].
\]
The term $\Phi^\top\Lambda z$ in $b$ comes from the same logit variation as the positive term $z^\top\Lambda z$. Retaining $\Phi^\top\Lambda\Phi$ in $H_A$ allows these terms to be combined by completing the square pointwise in $\alpha$:
\begin{equation}
\label{eq:app_lsi_square_completion}
\begin{aligned}
z^\top\Lambda z-b^\top H_A^{-1}b
=\min_u\bigl\{&\sigma^{-2}\norm u^2+2r^\top Eu\\
&+(\Phi u+z)^\top\Lambda(\Phi u+z)\bigr\}.
\end{aligned}
\end{equation}
The minimum is over head vectors $u\in\R^{(C-1)m}$. Expanding the expression gives $u^\top H_Au+2b^\top u+z^\top\Lambda z$, whose minimum is attained at $u=-H_A^{-1}b$. Dropping the last, nonnegative term in~\eqref{eq:app_lsi_square_completion} and minimizing the remaining quadratic gives
\begin{equation}
\label{eq:app_lsi_interaction_cost}
z^\top\Lambda z-b^\top H_A^{-1}b\ge-\sigma^2\norm{E^\top r}^2\ge-\frac{2\sigma^2nG}{m}.
\end{equation}
Thus the combined loss-curvature and variance terms leave a cost controlled by first feature derivatives, through $G$. The only remaining term is $\E_{\mu_W}[r^\top F\alpha]$, which contains second feature derivatives and requires different estimates at the two depths.

\paragraph{Two layers.}
Write $a_j$ and $a_{j,0}$ for column $j$ of $A$ and $A_0$, respectively. Gaussian integration by parts against the tilted head law gives
\begin{equation}
\label{eq:app_lsi_head_ibp}
\begin{aligned}
\E_{\mu_W}[r_i^\top a_j]
={}&\E_{\mu_W}[r_i^\top a_{j,0}]\\
&+\frac{\sigma^2}{\sqrt m}\E_{\mu_W}\Bigl[H_{1,j}(x_i)\operatorname{tr}\Lambda_i\\
&\hspace{25mm}-\sum_{t=1}^n H_{1,j}(x_t)r_i^\top r_t\Bigr].
\end{aligned}
\end{equation}
Indeed, for each coordinate of $a_j$, differentiating the loss gradient produces the corresponding diagonal entry of $H_{1,j}(x_i)\Lambda_i/\sqrt m$, while differentiating the likelihood density produces the final sum. Since $\norm{a_{j,0}}\le a_0$, Lemma~\ref{lem:app_lsi_head} yields
\begin{equation}
\label{eq:app_lsi_residual_head}
\left|\E_{\mu_W}[r_i^\top a_j]\right|\le2a_0+\frac{\sigma^2(1+2n)}{\sqrt m}.
\end{equation}
The prior mean supplies the first term in~\eqref{eq:app_lsi_residual_head}. The remaining terms gain the factor $m^{-1/2}$ from differentiating the logits, which is retained by estimating the product inside its expectation. For rows $W_{1,j}$ and $v_j$ of the weight and direction matrices, the coordinate formula is
\[
\bigl(D_W^2H_1(x_i)[v,v]\bigr)_j=\phi''(W_{1,j}x_i)(v_jx_i)^2.
\]
Keeping the product $r_i^\top a_j$ inside its expectation, and using $\sum_{i,j}(v_jx_i)^2\le\chi$, gives
\[
\left|\E_{\mu_W}[r^\top F\alpha]\right|\le c_\phi\chi\left(\frac{2a_0}{\sqrt m}+\frac{\sigma^2(1+2n)}m\right).
\]
Combining this with~\eqref{eq:app_lsi_interaction_cost} gives the shallow coefficients in~\eqref{eq:app_lsi_shallow_coefficients}.

\paragraph{Greater depth.}
For any fixed vector $z_i\in\R^m$, Cauchy--Schwarz and the projected head-moment estimate give
\[
\left|\E_{\mu_W}[r_i^\top Az_i]\right|\le\sqrt{2K}\norm{z_i}.
\]
Apply this with $z_i=D_W^2H_h(x_i)[v,v]$, which is fixed when conditioning on $W$, and use Lemma~\ref{lem:app_lsi_derivatives}. Including the output normalization gives
\[
\left|\E_{\mu_W}[r^\top F\alpha]\right|\le n\sqrt{2K}\left(\frac T{\sqrt m}+\frac{T_{\mathrm{mix}}}m\right).
\]
Together with~\eqref{eq:app_lsi_interaction_cost}, this gives~\eqref{eq:app_lsi_deep_coefficients}. The factor $\sqrt K=O(\sqrt n)$ in this projected-moment estimate accounts for the stronger sample-size dependence of the deep curvature loss. In both cases,~\eqref{eq:app_lsi_curvature_identity} is bounded below by $\rho$.
\end{proof}

\subsection{Conditional-score estimate}
\label{app:lsi_score}

The marginal LSI alone does not control how the head conditional changes with $W$. For a unit hidden direction $v$, the derivative of the log conditional density is the centered score
\begin{equation}
\label{eq:app_lsi_score_definition}
s_v(A)=\E_{\mu_W}[\partial_vV]-\partial_vV(A,W).
\end{equation}
Strong convexity supplies concentration for Lipschitz observables, but the head derivative of the score contains head-dependent terms. The next lemma retains both the Gaussian and likelihood parts of the conditional potential in a transport argument to control this dependence.

\begin{lemma}[Uniform score bound]
\label{lem:app_lsi_score}
For every $W$ in the relevant hidden domain, every unit hidden direction $v$, and every $t\in\R$,
\begin{equation}
\label{eq:app_lsi_score_mgf}
\log\E_{\mu_W}e^{t s_v}\le\frac{qt^2}{2},
\end{equation}
where $q$ is the explicit coefficient in~\eqref{eq:app_lsi_rho_q}.
\end{lemma}

\begin{proof}
Fix $W,v$ and write $E=E_v$, $\Phi=\Phi(W)$, and $T_v(\alpha)=\nabla\ell(\Phi\alpha)^\top E\alpha=\partial_vV$. Consider a probability law $\xi\ll\mu_W$ with finite relative entropy. The transport inequality of \citet[Proposition~1.1]{corderoerausquin2017transport}, applied to the conditional potential, supplies a coupling $\alpha\sim\mu_W$, $\alpha'\sim\xi$ such that
\begin{equation}
\label{eq:app_lsi_transport}
\operatorname{KL}(\xi\|\mu_W)\ge\E\left[\frac{\norm{\alpha'-\alpha}^2}{2\sigma^2}+B_\ell(\Phi\alpha',\Phi\alpha)\right].
\end{equation}
Here the excess of the loss above its tangent plane is
\[
B_\ell(z',z)=\ell(z')-\ell(z)-\nabla\ell(z)^\top(z'-z).
\]
Because $\ell$ is convex with Hessian at most $I/2$,
\begin{equation}
\label{eq:app_lsi_bregman_gradient}
B_\ell(z',z)\ge\norm{\nabla\ell(z')-\nabla\ell(z)}^2.
\end{equation}
For completeness, apply the descent inequality to the nonnegative convex function $u\mapsto B_\ell(u,z)$, whose minimum is zero at $z$: taking a step of twice its gradient from $z'$ proves~\eqref{eq:app_lsi_bregman_gradient}.

Set $r=\nabla\ell(\Phi\alpha)$ and $r'=\nabla\ell(\Phi\alpha')$. The difference to be controlled splits as
\[
T_v(\alpha')-T_v(\alpha)=r'^\top E(\alpha'-\alpha)+(r'-r)^\top E\alpha.
\]
The first term uses the bound $\norm{r'}^2\le2n$ together with the Gaussian displacement cost in~\eqref{eq:app_lsi_transport}. The second uses the loss-gradient cost in~\eqref{eq:app_lsi_bregman_gradient}. Its unprimed factor $E\alpha$ keeps the required second moment under $\mu_W$, where~\eqref{eq:app_lsi_map_derivative} applies, rather than under the arbitrary comparison law $\xi$. Cauchy--Schwarz, first in the coupling and then across these two contributions, gives
\begin{equation}
\label{eq:app_lsi_score_transport_bound}
\begin{aligned}
&\left|\E_\xi T_v-\E_{\mu_W}T_v\right|\\
&\quad\le\Bigl[2\operatorname{KL}(\xi\|\mu_W)\\
&\qquad\times\Bigl(2\sigma^2n\norm E_{\op}^2+\tfrac12\E_{\mu_W}\norm{E\alpha}^2\Bigr)\Bigr]^{1/2}\\
&\quad\le\sqrt{2q\operatorname{KL}(\xi\|\mu_W)}.
\end{aligned}
\end{equation}
The final step is~\eqref{eq:app_lsi_map_derivative}. For $\bar{x}_v=T_v-\E_{\mu_W}T_v$, the entropy variational formula now yields
\[
\begin{aligned}
\log\E_{\mu_W}e^{t\bar{x}_v}
&=\sup_{\xi\ll\mu_W}\{t\E_\xi \bar{x}_v-\operatorname{KL}(\xi\|\mu_W)\}\\
&\le\sup_{u\ge0}\{|t|\sqrt{2qu}-u\}=\frac{qt^2}{2}.
\end{aligned}
\]
All exponential moments exist because $|T_v(\alpha)|\le\sqrt{2n}\norm E_{\op}\norm\alpha$ and $\mu_W$ is a bounded-density tilt of a Gaussian. The transport step can first be applied to smooth densities and then extended by approximation. Since $s_v=-\bar{x}_v$, this proves the claim for both signs of $t$.
\end{proof}

\subsection{Entropy assembly}
\label{app:lsi_assembly}

We first convert marginal curvature into an LSI on the hidden domain, then use the score estimate to control differentiation of conditional second moments. The entropy argument is the conditional-score form of the two-scale framework of \citet[Theorem~3.1(3)]{srinivasan2026twoscale}; we give the calculation on our domain to specify the constant and the boundary treatment.

\begin{lemma}[LSI for the hidden marginal]
\label{lem:app_lsi_marginal}
If $\rho>0$, then $\CLS(\bar\nu)\le\rho^{-1}$, with ambient hidden-weight gradients on its support.
\end{lemma}

\begin{proof}
Lemma~\ref{lem:app_lsi_curvature} gives $\nabla_W^2\bar U\succeq\rho I$. For $L=2$, the usual Bakry--\'Emery criterion applies. For $L>2$, the hidden support is convex, since each operator-norm constraint is convex. The convex-domain criterion therefore gives the same constant \citep[Theorem~A1]{schlichting2019mixtures}.

One may justify the nonsmooth and unbounded domain directly by exhausting its interior with bounded smooth convex domains. On each such domain the potential retains its Hessian lower bound, and the reflecting Bakry--\'Emery argument has a nonnegative boundary term. The normalized restricted densities converge to $\bar\nu$; applying the inequalities to compactly supported smooth functions and passing to the limit preserves the constant $\rho^{-1}$. Cutoff and truncation approximation then extend the inequality to its finite-energy class. No boundary condition is imposed on the test functions in the resulting functional inequality.
\end{proof}

\begin{lemma}[Entropy assembly]
\label{lem:app_lsi_assembly}
If $\rho>0$, then $\CLS(\nu)\le\lambda_{\max}(M)$, with $M$ given by~\eqref{eq:app_lsi_matrix}.
\end{lemma}

\begin{proof}
For a smooth test function $g=g(A,W)$, set $F(W)=(\E_{\mu_W}g^2)^{1/2}$. Conditional entropy decomposition gives
\begin{equation}
\label{eq:app_lsi_entropy_split}
\Ent_\nu(g^2)=\E_{\bar\nu}\Ent_{\mu_W}(g^2)+\Ent_{\bar\nu}(F^2).
\end{equation}
At points where $F>0$, differentiation in a unit hidden direction gives
\begin{equation}
\label{eq:app_lsi_F_derivative}
\partial_vF=\frac{\E_{\mu_W}(g\partial_vg)}F+\frac{\E_{\mu_W}(g^2s_v)}{2F}.
\end{equation}
Apply the entropy variational inequality and~\eqref{eq:app_lsi_score_mgf} to the probability law with density $g^2/F^2$ relative to $\mu_W$. Optimizing the exponential parameter yields
\[
\left|\E_{\mu_W}(g^2s_v)\right|\le F\sqrt{2q\Ent_{\mu_W}(g^2)}.
\]
The conditional LSI from Lemma~\ref{lem:app_lsi_head}, followed by Cauchy--Schwarz in the first term of~\eqref{eq:app_lsi_F_derivative}, therefore gives
\begin{equation}
\label{eq:app_lsi_F_gradient}
\begin{aligned}
\norm{\nabla_WF}\le{}&(\E_{\mu_W}\norm{\nabla_Wg}^2)^{1/2}\\
&+\sigma\sqrt q\,(\E_{\mu_W}\norm{\nabla_Ag}^2)^{1/2}.
\end{aligned}
\end{equation}
Taking the supremum over unit directions introduces no dimension factor. Replacing $F$ by $(\E_{\mu_W}g^2+\tau)^{1/2}$ and sending $\tau\downarrow0$ handles its zeros.

Put $D_A=\E_\nu\norm{\nabla_Ag}^2$ and $D_W=\E_\nu\norm{\nabla_Wg}^2$. Apply the conditional and marginal LSIs to~\eqref{eq:app_lsi_entropy_split}, then use~\eqref{eq:app_lsi_F_gradient} and Cauchy--Schwarz over $W$:
\[
\begin{aligned}
\Ent_\nu(g^2)&\le2\sigma^2D_A+\frac2\rho\bigl(\sqrt{D_W}+\sigma\sqrt q\sqrt{D_A}\bigr)^2\\
&\le2\lambda_{\max}(M)(D_A+D_W).
\end{aligned}
\]
The first line is twice the quadratic form of $M$ evaluated at $(\sqrt{D_A},\sqrt{D_W})$. Its off-diagonal entry records the coupling introduced by the conditional score, and its largest eigenvalue bounds the cost relative to total gradient energy. This proves the stated LSI. The preceding calculation uses the ambient gradients on the product of the unrestricted head space and the hidden domain, so the convex restriction requires no further coupling or boundary term.
\end{proof}

To identify $\sigma^2$ as the limiting optimal constant, we also need a lower comparison. A scalar head coordinate supplies it: its likelihood curvature is at most $n/(2m)$, and it remains unrestricted under both $\pi$ and $\nu_a$.

\begin{lemma}[Matching lower comparison]
\label{lem:app_lsi_lower}
For either $\nu=\pi$ or $\nu=\nu_a$, independently of whether $\rho>0$,
\begin{equation}
\label{eq:app_lsi_lower_bound}
\CP(\nu)\ge\bigl(\sigma^{-2}+n/(2m)\bigr)^{-1}.
\end{equation}
\end{lemma}

\begin{proof}
Fix a scalar entry $z$ of the head $A$ and condition on all other parameters. This coordinate remains unrestricted. Its conditional potential $U_z$ has
\[
\sigma^{-2}\le U_z''\le\sigma^{-2}+\frac n{2m},
\]
because each hidden feature coordinate has magnitude at most one and the loss Hessian is at most $I/2$. Write $\E_z$, $\Var_z$, and $\Cov_z$ for this one-dimensional conditional law. Integration by parts gives
\[
\E_zU_z'=0,\qquad \Cov_z(z,U_z')=1,\qquad \E_z(U_z')^2=\E_zU_z''.
\]
The Gaussian tails justify the boundary terms. Hence Cauchy--Schwarz yields $\Var_z(z)\ge(\sigma^{-2}+n/(2m))^{-1}$. Total variance gives the same lower bound for $\Var_\nu(z)$. Testing the PI with this coordinate, whose gradient has norm one, proves the claim; smooth cutoffs justify its use if the PI is initially defined on compactly supported functions.
\end{proof}

\begin{proof}[Proof of Theorem~\ref{thm:app_lsi_finite}]
Lemmas~\ref{lem:app_lsi_marginal} and~\ref{lem:app_lsi_assembly} give the upper bound, while Lemma~\ref{lem:app_lsi_lower} gives the lower bound. Linearizing the LSI gives $\CP(\nu)\le\CLS(\nu)$. Solving $\kappa_1/\sqrt m+\kappa_2/m<\sigma^{-2}$ as a quadratic inequality in $\sqrt m$ proves~\eqref{eq:app_lsi_positive_width}.

Under~\eqref{eq:app_lsi_half_width}, the two terms in $\delta_m$ are each at most $(4\sigma^2)^{-1}$. Thus $\rho\ge(2\sigma^2)^{-1}$ and
\[
\rho^{-1}-\sigma^2=\frac{\sigma^2\delta_m}{\rho}\le2\sigma^4\delta_m.
\]
The largest eigenvalue of a symmetric two-by-two matrix is at most its larger diagonal entry plus the absolute off-diagonal entry. Applied to~\eqref{eq:app_lsi_matrix}, this gives
\[
\lambda_{\max}(M)\le\sigma^2+2\sigma^4\delta_m+2\sigma^4q+2\sigma^3\sqrt q,
\]
which is~\eqref{eq:app_lsi_additive_bound}.

The explicit coefficients now give $\delta_m=O(n/\sqrt m+n^2/m)$ for two layers and $\delta_m=O_{L,a}(n^{3/2}/\sqrt m+n^2/m)$ at greater depth, with $q=O_{L,a}(n^2/m)$ in both cases. This proves the main-text rates. If $m/n^2\to\infty$ in the shallow case, or $m/n^3\to\infty$ in the deep case, the upper bound tends to $\sigma^2$. Since $n/m\to0$ along either regime, the lower bound gives convergence of both optimal constants to $\sigma^2$.
\end{proof}

\begin{proof}[Proof of Corollary~\ref{cor:app_lsi_accuracy}]
The three nontrivial width requirements in~\eqref{eq:app_lsi_accuracy_width} give
\[
\delta_m\le\frac{\varepsilon}{8\sigma^2},\qquad q\le\frac{\varepsilon^2}{16\sigma^2}.
\]
In particular $\rho\ge(2\sigma^2)^{-1}$, so the argument giving~\eqref{eq:app_lsi_additive_bound} applies. Substitution yields
\[
\CLS(\nu)-\sigma^2\le\sigma^2\left(\frac\varepsilon4+\frac{\varepsilon^2}8+\frac\varepsilon2\right)\le\varepsilon\sigma^2.
\]
The mass statement follows from Proposition~\ref{prop:app_lsi_mass}.
\end{proof}

\subsection{Refinement using input geometry}
\label{app:lsi_input_geometry}

The shallow coefficients retain more information about the inputs than their individual norm bound. Recall the quantity in~\eqref{eq:app_lsi_chi}:
\[
\chi=\lambda_{\max}\!\left(\sum_{i=1}^n x_ix_i^\top\right)=\sup_{\norm u=1}\sum_{i=1}^n(u^\top x_i)^2.
\]
It measures the largest aggregate squared projection of the inputs onto one direction. The bound $\chi\le n\bar{x}^2$ gives the sample-size dependence in the main theorem; retaining $\chi$ gives the following refinement.

\begin{corollary}[Input-dependent shallow LSI]
\label{cor:app_lsi_input_geometry}
Consider the posterior in~\eqref{eq:app_lsi_posterior} with $L=2$, under~\eqref{eq:app_lsi_assumptions}. If
\begin{equation}
\label{eq:app_lsi_input_width}
\begin{split}
m\ge\max\bigl\{&1,64\sigma^4c_\phi^2a_0^2\chi^2,\\
&4\sigma^4\chi[c_\phi(1+2n)+2n]\bigr\},
\end{split}
\end{equation}
then
\begin{equation}
\label{eq:app_lsi_input_rate}
\CLS(\pi)\le\sigma^2+O\!\left(\sqrt{\frac{n\chi}{m}}+\frac{n\chi}{m}\right).
\end{equation}
The implicit constant depends only on $C,\sigma,\bar{x},c_\phi,a_0$ and is independent of $n,m,d$. In particular, for orthonormal inputs, $\chi=1$ and both $\CP(\pi)$ and $\CLS(\pi)$ converge to $\sigma^2$ when $m \gg n$. The first-layer prior center remains arbitrary.
\end{corollary}

\begin{proof}
Substituting~\eqref{eq:app_lsi_shallow_coefficients} into~\eqref{eq:app_lsi_half_width} gives exactly~\eqref{eq:app_lsi_input_width}. Thus the finite-width bound~\eqref{eq:app_lsi_additive_bound} applies, with
\[
\begin{aligned}
\delta_m&=\frac{2c_\phi a_0\chi}{\sqrt m}+\frac{\sigma^2\chi[c_\phi(1+2n)+2n]}{m},\\
q&=\frac\chi m\left(2\sigma^2n+\frac K2\right).
\end{aligned}
\]
The definition of $K$ in~\eqref{eq:app_lsi_head_constants} gives $K=O(n)$ with the stated dependence of constants. Also, $\chi\le n\bar{x}^2$ implies
\[
\frac\chi{\sqrt m}\le \bar{x}\sqrt{\frac{n\chi}{m}}.
\]
Consequently $\delta_m=O(\sqrt{n\chi/m}+n\chi/m)$ and $q=O(n\chi/m)$. Equation~\eqref{eq:app_lsi_additive_bound} now proves~\eqref{eq:app_lsi_input_rate}. Since $\chi^2\le n\bar{x}^2\chi$, the width threshold in~\eqref{eq:app_lsi_input_width} is at most a constant multiple of $\max\{1,n\chi\}$.

For orthonormal inputs, the matrix $\sum_i x_ix_i^\top$ is an orthogonal projection, so $\chi=1$. The width condition then holds eventually when $m/n\to\infty$, and the upper bound tends to $\sigma^2$. The lower comparison in Lemma~\ref{lem:app_lsi_lower} tends to the same value because $n/m\to0$, proving both limits.
\end{proof}

\subsection{Posterior mass of the spectral domain}
\label{app:spectral_mass}

This estimate uses the bounded activation and prior-center bounds from~\eqref{eq:app_lsi_assumptions}; it does not require the input bound or a positive marginal-curvature estimate.

\begin{proposition}[Explicit posterior-mass bound]
\label{prop:app_lsi_mass}
For $L>2$ and $a>r_0+2\sigma$, let $c_0$ be as in~\eqref{eq:app_lsi_head_constants} and set
\[
c_1=\frac{(a-r_0-2\sigma)^2}{2\sigma^2}>0.
\]
Then
\begin{equation}
\label{eq:app_lsi_mass}
\pi(\mathcal G_a^c)\le(L-2)e^{nc_0-c_1m}.
\end{equation}
Consequently, for every $0<\eta<1$, the condition
\begin{equation}
\label{eq:app_lsi_mass_width}
m\ge\frac{2\sigma^2}{(a-r_0-2\sigma)^2}\left[nc_0+\log\frac{L-2}{\eta}\right]
\end{equation}
ensures $\pi(\mathcal G_a)\ge1-\eta$.
\end{proposition}

\begin{proof}
Averaging the conditional evidence estimate in Lemma~\ref{lem:app_lsi_head} over the hidden Gaussian prior gives $Z\ge e^{-nc_0}$. As $V\ge0$, every measurable event $B$ satisfies
\[
\pi(B)=Z^{-1}\E_\gamma[e^{-V}\mathbf1_B]\le e^{nc_0}\gamma(B).
\]
For a standard Gaussian $m\times m$ matrix $G_\ell$, the expected operator norm is at most $2\sqrt m$, and Gaussian concentration in the Frobenius metric gives
\[
\Pr\{\norm{G_\ell}_{\op}>2\sqrt m+t\}\le e^{-t^2/2},\qquad t\ge0;
\]
see \citet[Theorem~5.32 and Proposition~5.34]{vershynin2012introduction}. Under the prior, write $W_\ell=W_{\ell,0}+\sigma G_\ell$. The prior-center bound and the triangle inequality yield
\[
\gamma\{\norm{W_\ell/\sqrt m}_{\op}>a\}\le e^{-c_1m}.
\]
A union bound over the $h-1=L-2$ restricted matrices, followed by posterior domination, proves~\eqref{eq:app_lsi_mass}. Solving its right-hand side for a value at most $\eta$ gives~\eqref{eq:app_lsi_mass_width}.
\end{proof}

\subsection{Depth dependence}
\label{app:lsi_depth}

The finite-width formulas expose all dependence on depth and the cutoff. Suppose $a=1+\varepsilon$ with $\varepsilon\ge0$, and put $\bar{x}_*=\max\{1,\bar{x}\}$. From~\eqref{eq:app_lsi_p_vectors}--\eqref{eq:app_lsi_derivative_constants},
\begin{equation}
\label{eq:app_lsi_depth_constants}
\begin{aligned}
S&\le \bar{x}_*\sqrt h\,e^{\varepsilon h},\\
T&\le\tfrac12c_\phi \bar{x}_*^2h(h+1)e^{2\varepsilon h},\\
T_{\mathrm{mix}}&\le \bar{x}_*(h-1)e^{\varepsilon h}.
\end{aligned}
\end{equation}
Indeed, $\norm{p_\ell}^2\le \bar{x}_*^2\ell e^{2\varepsilon(\ell-1)}$. Bounding the largest eigenvalue in $T$ by the trace and summing over $\ell$ gives the second line. Each off-diagonal entry of $R$ is at most $\bar{x}_*e^{\varepsilon h}$, so its maximal row sum gives the third.

For a specified LSI accuracy and mass, it is enough to take the maximum of the right-hand sides of~\eqref{eq:app_lsi_accuracy_width} and~\eqref{eq:app_lsi_mass_width}, with the deep coefficients in~\eqref{eq:app_lsi_deep_coefficients}. In particular, the contribution of $\kappa_1^2=2n^2KT^2$ has order $h^4e^{4\varepsilon h}n^3$ at fixed accuracy and model scales. The remaining terms are explicitly retained in those width conditions. When $\varepsilon h$ is bounded, this leading contribution is of order $h^4n^3$.

The cutoff must also satisfy $1+\varepsilon>r_0+2\sigma$. With centered later-layer priors and $\sigma=1/2$, any $\varepsilon>0$ is admissible, and~\eqref{eq:app_lsi_mass_width} becomes
\[
m\ge\frac{nc_0+\log((L-2)/\eta)}{2\varepsilon^2}.
\]
Thus reducing the cutoff slack limits derivative growth through depth while increasing the width needed for a prescribed posterior mass. These are simultaneous requirements on the single domain $\mathcal G_a$.

\section{Additional experimental results}
\label{app:experiments}
\subsection{Model, sampling, and estimators}
\label{app:experiments_setup}

\paragraph{Model and data.}
We consider binary classification with input dimension $d=32$, $n=128$ training observations, and $1{,}024$ held-out observations. Each replicate draws a teacher vector $v\sim N(0,I_{32})$. Independently generated inputs have the form $x=z/\norm z$, where $z\sim N(0,I_{32})$, and labels follow
\[
\Pr(y=1\mid x,v)=\frac{1}{1+e^{-2x^\top v}}.
\]
The training and held-out sets are independent conditional on the teacher. Within each of three replicates, the same teacher and data are used for all widths and both architectures; the teacher and data are redrawn across replicates.

The networks are bias-free and use $\tanh$ activations. With the normalization in~\eqref{eq:network}, $H_1(x)=\tanh(W_1x)$, the deep model adds $H_2(x)=\tanh(W_2H_1(x)/\sqrt m)$, and both use $f_\theta(x)=AH_{L-1}(x)/\sqrt m$. The shallow widths are $64,256,1024,4096$ at $L=2$; the deep widths are $32,64,128,256$ at $L=3$. We encode the two classes as $0$ and $1$, corresponding to classes $1$ and $2$ in Section~\ref{sec:model}. The centered logits $(-f_\theta/\sqrt2,f_\theta/\sqrt2)$ then give the class-$1$ probability in this binary coding
\[
p_\theta(x)=\frac{1}{1+\exp(-\sqrt2 f_\theta(x))}.
\]
The loss $V$ in~\eqref{eq:loss} is their summed binary cross-entropy. All calculations use double precision.

The prior and posterior are those of~\eqref{eq:posterior}, with $\sigma=1$ over all parameter coordinates. The head center is deterministic: $(A_0)_j=(-1)^{j-1}/\sqrt m$ for $1\le j\le m$, giving $\norm{A_0}_{\op}=1$. Each replicate draws a $4096\times32$ bank of independent standard Gaussian entries; $W_{1,0}$ uses its first $m$ rows, shared across architectures. The deep model has $W_{2,0}=0$. Thus $\tanh$ satisfies Assumption~\ref{ass:activation}, the centers satisfy Assumption~\ref{ass:prior_means} with $a_0=1$ and $r_0=0$, and Assumption~\ref{ass:inputs} holds with $\bar{x}=1$. For $L=3$, the domain in~\eqref{eq:lsi_domain} restricts $W_2$ alone, with admissible fixed cutoffs $a>r_0+2\sigma=2$ for the deep LSI.

\paragraph{Reference samples.}
Full-state elliptical slice sampling \citep{murray2010elliptical}, centered at $\theta_0$, provides the reference samples for spectral coverage and predictive scores. For each target (a combination of replicate, architecture, and width), four chains start from independent prior draws and use independent random streams. Each chain discards $7{,}168$ transitions: $4{,}096$ for burn-in, $2{,}048$ for probe calibration, and $1{,}024$ for separation before production. We then run $65{,}536$ production transitions and archive every sixteenth state, giving $4{,}096$ states per chain and $16{,}384$ per target. Both the reference sampler and the Langevin sampler below update all parameter blocks and target the unrestricted posterior.

\paragraph{Langevin traces.}
We use a Metropolis-adjusted Gaussian-reference Langevin proposal \citep[Section~4.3]{cotter2013mcmc}. Parameterize the step by $h>0$ and set $\eta=e^{-h/\sigma^2}$. For a standard Gaussian vector $\xi$ in parameter space, propose
\begin{equation}
\label{eq:experimental_langevin_proposal}
\begin{aligned}
\theta'={}&\theta_0+\eta(\theta-\theta_0)\\
&-\sigma^2(1-\eta)\nabla V(\theta)\\
&+\sigma\sqrt{1-\eta^2}\,\xi.
\end{aligned}
\end{equation}
This integrates the Gaussian prior drift exactly while holding the likelihood gradient fixed. Let $q_h(\theta'\mid\theta)$ be this Gaussian proposal density and $U(\theta)=V(\theta)+\norm{\theta-\theta_0}^2/(2\sigma^2)$. The Metropolis--Hastings acceptance probability \citep{hastings1970monte} is
\[
\min\!\left\{1,\ e^{U(\theta)-U(\theta')}\frac{q_h(\theta\mid\theta')}{q_h(\theta'\mid\theta)}\right\}.
\]
Gradients use the full summed loss. The step is fixed throughout each trajectory, and rejected proposals remain as repeated states.

Each Langevin chain starts from the final production state of its corresponding reference chain and discards an accumulated step parameter of $20$: $2{,}000$ transitions at $h=0.01$ or $4{,}000$ at $h=0.005$. The main loss and prediction analyses use $102{,}400$ consecutive retained transitions per chain at the common step $h=0.01$. The step-halving comparison uses the deep endpoints $m=32,256$ and $N$ retained transitions chosen so that $Nh=512$: $51{,}200$ transitions at $h=0.01$ and $102{,}400$ at $h=0.005$. Both steps start from the same four reference end states, with independent noise streams. Here $h$ is the prior Ornstein--Uhlenbeck time parameter in~\eqref{eq:experimental_langevin_proposal}, and $kh$ is the algorithmic lag after $k$ transitions.

\paragraph{Estimators.}
For the deep model, $T=\norm{W_2}_{\op}/\sqrt m$. Within each target, empirical coverage is the fraction of its $16{,}384$ archived states satisfying $T\le a$; quantiles use linear interpolation. Prior spectral comparisons use $4{,}096$ independent standard Gaussian matrices at each width, with the same prior reference shared across replicates. Points and whiskers report medians and ranges of replicate-specific estimates.

For a scalar sequence $z_0,\ldots,z_{N-1}$ with sample mean $\bar z$, the chain autocorrelation at lag $k$ is
\begin{equation}
\label{eq:experimental_acf}
\widehat\rho(k)=\frac{\sum_{t=0}^{N-k-1}(z_t-\bar z)(z_{t+k}-\bar z)}{\sum_{t=0}^{N-1}(z_t-\bar z)^2}.
\end{equation}
We use consecutive retained transitions and average the four separately normalized chain autocorrelations within each replicate. For predictions, we additionally average over eight fixed held-out inputs, selected before sampling at zero-based indices $0,128,256,384,512,640,768,896$. These inputs are shared across widths and architectures within the replicate. Lines and shading show the median and range across the three replicates, which are the experimental replication units.

\subsection{Coverage of the spectral domain}
\label{app:experimental_coverage}

This analysis uses the $L=3$ model from Section~\ref{app:experiments_setup}. It measures the mass of the domain in~\eqref{eq:lsi_domain}, on which Theorem~\ref{thm:deep_lsi} proves an LSI. Lemma~\ref{lem:spectral_mass} controls this mass under Assumptions~\ref{ass:activation} and~\ref{ass:prior_means}.

\begin{table*}[t]
\centering
\caption{\textbf{Empirical coverage of the deep spectral domain.} Percentages of unrestricted reference states satisfying $T\le a$, reported as median [minimum, maximum] across three replicates, each with $16{,}384$ inspected states from four chains. The column $a=2$ is the threshold for admissibility; $2.05$ and $2.10$ are illustrative fixed cutoffs above it. At $m=256$ and $a=2.10$, $100.00\%$ records zero observed exceedances.}
\label{tab:experimental_coverage}
\begin{tabular}{rccc}
\toprule
Width $m$ & $a=2$ & $a=2.05$ & $a=2.10$ \\
\midrule
32  & 86.47 [85.13, 86.99] & 95.37 [94.65, 95.43] & 98.66 [98.36, 98.68] \\
64  & 86.34 [85.36, 86.54] & 97.62 [97.26, 97.65] & 99.74 [99.68, 99.77] \\
128 & 85.80 [85.78, 85.95] & 99.19 [99.19, 99.22] & 99.98 [99.96, 99.99] \\
256 & 85.78 [85.41, 85.85] & 99.92 [99.89, 99.93] & 100.00 [100.00, 100.00] \\
\bottomrule
\end{tabular}
\end{table*}

\begin{rem}[Coverage at the spectral threshold]
\label{rem:experimental_cutoff}
Table~\ref{tab:experimental_coverage} shows increasing coverage at the fixed cutoffs $a=2.05$ and $2.10$, while coverage at $a=2$ stays near $86\%$. This is compatible with the upper quantiles approaching $2$: a distribution can concentrate increasingly tightly around a threshold while retaining mass just above it. Lemma~\ref{lem:spectral_mass} concerns each fixed cutoff strictly above $2$, where that near-threshold mass is eventually included. The empirical comparison therefore distinguishes concentration near the threshold from coverage at the threshold itself.
\end{rem}

The cutoffs here belong to the deep LSI domain. The larger auxiliary cutoff in Theorem~\ref{thm:global_pi} serves the separate restriction-removal argument for the global PI.

\begin{table}[t]
\centering
\caption{\textbf{Prior and posterior spectral quantiles.} Median and 99th percentile of $T$ for the deep model. Posterior entries are medians of three replicate-specific quantiles; prior entries use $4{,}096$ independent Gaussian matrices at each width. The prior reference is shared across replicates.}
\label{tab:prior_spectral_comparison}
\begin{tabular}{rcccc}
\toprule
& \multicolumn{2}{c}{Median} & \multicolumn{2}{c}{99th percentile} \\
\cmidrule(lr){2-3}\cmidrule(lr){4-5}
$m$ & Posterior & Prior & Posterior & Prior \\
\midrule
32  & 1.908 & 1.905 & 2.112 & 2.111 \\
64  & 1.943 & 1.944 & 2.073 & 2.075 \\
128 & 1.965 & 1.965 & 2.046 & 2.050 \\
256 & 1.978 & 1.978 & 2.029 & 2.028 \\
\bottomrule
\end{tabular}
\end{table}

Table~\ref{tab:prior_spectral_comparison} shows that the prior and posterior quantiles of this statistic are close. The predictive comparison below examines how conditioning on the data changes predictions despite this similar hidden-weight spectral scale.

\subsection{Loss and held-out predictions}
\label{app:experimental_predictions}

\begin{figure}[t]
\centering
\includegraphics[width=0.6\linewidth]{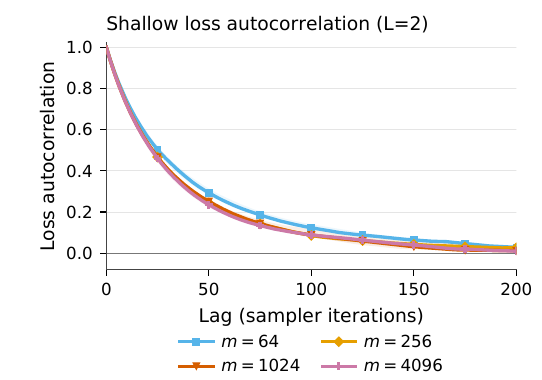}
\caption{\textbf{Loss autocorrelation for shallow networks ($L=2$).} Autocorrelation of summed training cross-entropy at widths $m=64,256,1024,4096$ under the Metropolis-adjusted Gaussian-reference Langevin sampler. The sampler targets the unrestricted posterior at step size $0.01$, initialized from the reference chains, and uses $102{,}400$ retained transitions per chain. Autocorrelations are computed separately for four chains and averaged within replicate; curves show medians across three replicates. The data and prior construction are shared with the $L=3$ comparison in Figure~\ref{fig:experiments}(c); Section~\ref{app:experiments_setup} gives the full setup.}
\label{fig:shallow_loss_acf}
\end{figure}

These measurements use the shallow ($L=2$) and deep ($L=3$) models from Section~\ref{app:experiments_setup}, satisfying Assumptions~\ref{ass:activation}--\ref{ass:inputs}. Figure~\ref{fig:shallow_loss_acf} gives the shallow counterpart to Figure~\ref{fig:experiments}(c). Over the tested width range, the shallow model shows smaller reductions in loss autocorrelation, with little further change at the largest widths.

We next examine whether the width effect extends from training loss to predictions. The prediction observable is $p_\theta(x)$ at each of the eight fixed held-out inputs.

\begin{figure*}[t]
\centering
\includegraphics[width=\textwidth]{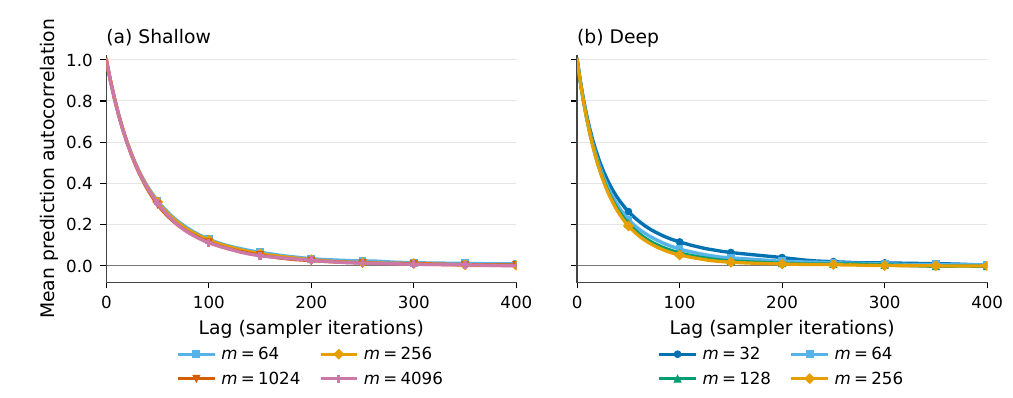}
\caption{\textbf{Autocorrelation of held-out predictions.} Class-$1$ prediction probabilities at eight fixed held-out inputs for shallow ($L=2$, left) and deep ($L=3$, right) networks. Autocorrelation is computed separately for each chain and input, then averaged within replicate. Lines and shading show the median and range over three replicates. The model, step size $0.01$, and $102{,}400$-transition windows are the same as in the main loss analysis; see Section~\ref{app:experiments_setup}.}
\label{fig:prediction_acf}
\end{figure*}

The deep width effect extends to predictions: the widest endpoint has lower mean prediction autocorrelation at every displayed nonzero lag in each of the three replicates. At the inspected lags from $25$ to $100$, the wider endpoint also has lower autocorrelation at each of the eight inputs in every replicate. Shallow gains are smaller.

To assess predictive quality, we use $2{,}048$ posterior parameter draws per target: every eighth archived reference state starting from the first, giving $512$ draws per chain at spacing $128$ production transitions. The prior predictive mixture uses $2{,}048$ independent draws from that target's prior, generated separately for every replicate, architecture, and width. For each held-out observation, we average the probability assigned to its label over these draws before taking the negative logarithm, then average over all $1{,}024$ observations. Let $\Delta\mathrm{NLS}$ be this score for the prior predictive distribution minus the score for the posterior predictive distribution, so positive values favor the posterior.

\begin{table}[t]
\centering
\caption{\textbf{Held-out predictive improvement.} Reduction $\Delta\mathrm{NLS}$ in mean negative log predictive probability, in nats, on $1{,}024$ held-out observations. Entries show median [minimum, maximum] across three replicates; positive values favor the posterior. Predictive mixtures use $2{,}048$ parameter draws per target.}
\label{tab:predictive_scores}
\begin{tabular}{lrl}
\toprule
Model & $m$ & $\Delta\mathrm{NLS}$ \\
\midrule
Shallow & 64   & 0.112 [0.071, 0.144] \\
Shallow & 256  & 0.118 [0.081, 0.136] \\
Shallow & 1024 & 0.114 [0.077, 0.138] \\
Shallow & 4096 & 0.114 [0.083, 0.139] \\
Deep    & 32   & 0.091 [0.059, 0.124] \\
Deep    & 64   & 0.088 [0.059, 0.123] \\
Deep    & 128  & 0.091 [0.055, 0.113] \\
Deep    & 256  & 0.095 [0.061, 0.108] \\
\bottomrule
\end{tabular}
\end{table}

Posterior predictive scores improve over the prior in every target. The improvements remain comparable across widths, providing predictive context for the decorrelation comparison.

\subsection{Step-size sensitivity}
\label{app:experimental_step_size}

This comparison uses the same $L=3$ model, data, and priors as the deep main experiment, retaining Assumptions~\ref{ass:activation}--\ref{ass:inputs}. It compares widths $m=32$ and $256$ at steps $0.01$ and $0.005$, with equal retained accumulated step parameter $Nh=512$ per chain.

\begin{figure*}[t]
\centering
\includegraphics[width=\textwidth]{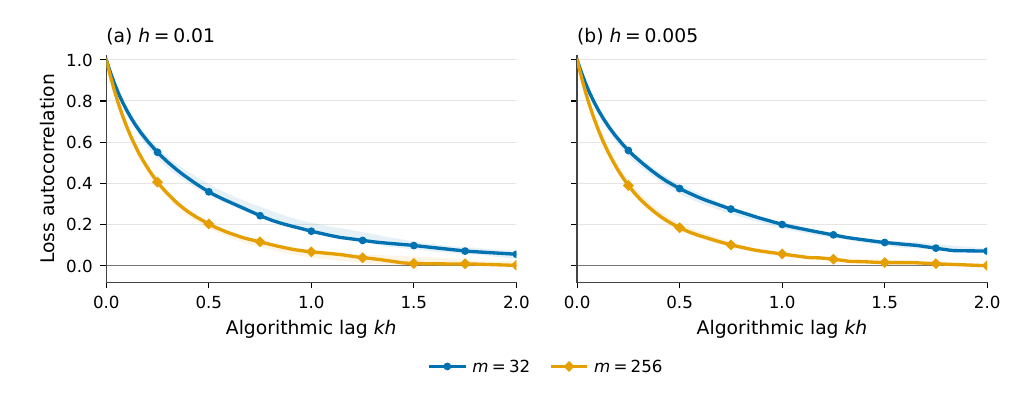}
\caption{\textbf{Deep width comparison under step halving.} Training-loss autocorrelation at widths $32$ and $256$ for step size $h=0.01$ (left) and $h=0.005$ (right). The horizontal axis is algorithmic lag $kh$, where $k$ counts transitions and $h$ is the step defined in~\eqref{eq:experimental_langevin_proposal}. Both analyses use retained duration $Nh=512$ per chain. Curves average four separately normalized chain autocorrelations; lines and shading show the median and range across three replicates. Model and analysis details are in Section~\ref{app:experiments_setup}.}
\label{fig:step_size}
\end{figure*}

The wider endpoint has lower loss autocorrelation at both step sizes, at every displayed nonzero algorithmic lag through $2$ in each replicate. The endpoint ordering therefore survives this step-size change for the finite-lag behavior of the tested sampler.

\subsection{Sampling diagnostics}
\label{app:experimental_diagnostics}

We summarize rank-normalized split $\widehat R$ and bulk and tail effective sample sizes following \citet{vehtari2021rank}. We report the larger of the bulk and folded split $\widehat R$ values. Bulk effective sample size uses Geyer's initial-positive-monotone estimator on rank-normalized split chains; tail effective sample size is the smaller value for the $5\%$ and $95\%$ quantile indicators. All series use four chains. The analysis criteria are $\widehat R<1.01$, bulk effective sample size at least $1{,}000$, and tail effective sample size at least $400$.

The reference pool contains $24$ targets, with $26$ monitored observables per shallow target and $28$ per deep target. These include $V$ and $U$, eight training logits, the eight held-out probabilities, four head--hidden interaction probes, two fixed parameter projections, and mean squared displacements from the prior centers for $A$ and $W_1$. Deep targets additionally monitor the displacement of $W_2$ and its spectral statistic. Diagnostics use the $65{,}536$ production transitions per chain, except for the spectral statistic, which uses the $4{,}096$ archived states. The dynamics pool contains loss and eight predictions for all $24$ targets; the step-size pool contains loss for two deep widths, two steps, and three replicates.

\begin{table*}[t]
\centering
\caption{\textbf{Diagnostics for the reported analyses.} Worst values across monitored target--observable series: largest rank-normalized split $\widehat R$ and smallest bulk and tail effective sample sizes (ESS). ESS values are rounded down. The spectral row is a subset of the reference pool, and monitored-series counts are not counts of independent experiments. Each series uses four chains.}
\label{tab:experimental_diagnostics}
\begin{tabular}{lrrrr}
\toprule
Analysis pool & Series & Max $\widehat R$ & Min bulk ESS & Min tail ESS \\
\midrule
Reference samples, all observables & 648 & 1.00513 & 1,116 & 2,376 \\
Reference spectral statistic (subset) & 12 & 1.00069 & 5,172 & 8,512 \\
Main/prediction dynamics: loss and predictions & 216 & 1.00280 & 3,212 & 7,572 \\
Step-size comparison: loss & 12 & 1.00423 & 1,429 & 3,308 \\
\bottomrule
\end{tabular}
\end{table*}

All monitored series satisfy the stated criteria. Reference and dynamics diagnostics are reported separately because spectral estimates and autocorrelation curves use different sample pools. These checks support the reported observable estimates; the width comparison concerns their finite-lag decorrelation under the specified discrete sampler.
\end{document}